\documentclass[lettersize,journal]{IEEEtran}
\usepackage{amsmath,amsfonts}
\usepackage{algorithmic}
\usepackage{algorithm}
\usepackage{array}
\usepackage[caption=false,font=footnotesize,labelfont=sf,textfont=sf]{subfig}
\usepackage{textcomp}
\usepackage{stfloats}
\usepackage{url}
\usepackage{verbatim}
\usepackage{graphicx}
\usepackage{cite}
\usepackage{booktabs}
\usepackage{amsmath}
\usepackage{amssymb}
\usepackage{tabularx}
\usepackage{multirow}
\usepackage{bbm}
\usepackage[framemethod=tikz]{mdframed}

\begin{document}

\title{Can Large Language Models Explain Flight Safety Events? A Prior-Guided Semantic LLM-based Approach}

\author{
    Lu~Xu,
    ~Xu~Li,
    ~Linjiang~Zheng,~\IEEEmembership{Member,~IEEE,}
    ~Fan~Li,
    ~Riquan~Zhang,
    ~Jiaxing~Shang,~\IEEEmembership{Member,~IEEE}
\thanks{Manuscript received xx xx, 2026; revised xx xx, 2026. This work was supported in part by the following: the Sichuan Science and Technology Program (No. 2025YFHZ0025), the Research Project of Sichuan Flight Engineering Technology Research Center (No. GY2025-04B), the National Natural Science Foundation of China (Nos. 12531013, 12371272).}
\thanks{L. Xu, X. Li, L. Zheng, J. Shang are with College of Computer Science, Chongqing University, Chongqing, China. (e-mail: xulu\_yzyz@163.com; leuio@foxmail.com; zlj\_cqu@cqu.edu.cn; shangjx@cqu.edu.cn.)}
\thanks{F. Li is with Sichuan Flight Engineering Technology Research Center, Civil Aviation Flight University of China, Guanghan, China. E-mail: lifan@cafuc.edu.cn}
\thanks{R. Zhang is with School of Statistics and Data Science, Shanghai University of International Business and Economics, Shanghai, China. E-mail: zhangriquan@163.com}
\thanks{*Corresponding author: Jiaxing Shang}}

\markboth{IEEE Transactions on Intelligent Transportation Systems}%
{Xu \MakeLowercase{\textit{et al.}}: A Sample Article Using IEEEtran.cls for IEEE Journals}

\maketitle

\begin{abstract}
Improving flight safety with multivariate time series flight data requires not only accurate detection of risk events, but more importantly, clear interpretation of their underlying causes at the level of pilot control behavior. Existing explainable AI techniques, such as feature importance maps, often require considerable domain knowledge to translate them into operationally meaningful explanations. Large Language Models (LLMs), which excel at language reasoning, bring a promising solution to this issue. However, applying LLMs in this domain presents several key challenges, including modal inconsistency, limited classification ability, scarcity of task-specific data for fine-tuning, and lack of domain knowledge. 
To overcome these challenges, we propose FlightLLM, a prior-guided semantic LLM-based approach for interpretable flight safety analysis. Specifically, we first perform feature engineering to address modal inconsistency, combining statistical descriptors with physically meaningful flight indicators. This representation is further processed by a Semantic Discretization module, which converts abstract numerical patterns into qualitative descriptions that are more compatible with language reasoning. 
In addition, since LLMs are not inherently strong classifiers, CatBoost is incorporated as a statistical expert, and its prediction results are injected into the prompt as prior guidance. 
A contrastive few-shot learning strategy is further adopted to compensate for limited data. Finally, we design structured prompts to embed aviation-specific knowledge into the inference process. Using hard landing, a representative risk event with complex causal mechanisms, as an anchor point, we evaluate FlightLLM on a dataset of 704 real-world A320 flight samples. Experimental results demonstrate that the proposed approach achieves competitive classification performance while generating direct and reasonable explanations for event causes. This work provides a new solution to predict and explain flight safety events.
\end{abstract}


\section{Introduction}
\IEEEPARstart{C}{ivil} aviation is widely recognized as one of the safest ways of transportation. However, the prevention and control of flight safety incidents remains a core focus of the industry. According to safety reports from the International Air Transport Association (IATA), taking measures to reduce operational risk occurrences remains a top priority. Consequently, conducting detailed analysis on the process of flight safety events is of significance for reducing incident rates and enhancing the level of aviation safety. In modern civil aviation field, ensuring flight safety has evolved from simple event monitoring to analysis of complex factors. The objective is no longer limited to identifying what happened, but to uncovering the underlying causal mechanisms, particularly those related to pilot control behavior and interactions with the environment. For experts and pilots, understanding the reasons behind the incidents is essential for designing effective training programs and risk mitigation strategies. 

To explore these causes from an actionable angle and mitigate future risks, the aviation industry uses Quick Access Recorder (QAR), an onboard device which collects up to 2000 flight parameters, to provide the foundational data for research. QAR data consists of time series with multiple variables that record the entire flight cycle, from engine start to shutdown. It captures diverse information, including equipment status (e.g., landing gear, flaps), pilot control inputs (e.g., pitch, roll), and kinematic parameters (e.g., altitude, vertical speed).
Researchers leverage QAR data to investigate various flight risk events, such as tail strike~\cite{wangDiscover}, airport risk~\cite{WangReseach}, hard landing~\cite{LiCurve}, long landing~\cite{Yang2022DataDriven}, and runway veer-off~\cite{MorettiRunway}. 
Early studies often relied on parameter exceedance signals~\cite{xiangzhang2022risk} to assess risks, which overlooks the potential relationships between parameters. Traditional machine learning methods such as clustering algorithms~\cite{LiCurve} and SVM~\cite{Hu2016SVM} have been applied to classify and analyze hard landing. Although effective in classification and prediction, these methods prioritized detection metrics over causal explanations. They rarely answered why an incident occurred.
With the continuous development of deep learning, modern approaches like LSTM~\cite{tong2018innovative} and Transformer~\cite{shang2025dual,chen2023sdtan} have also been used in this field. These approaches excel at processing high-dimensional time-series data and achieve high accuracy.
Notably, prior studies based on Transformer architectures have attempted to enhance interpretability by visualizing attention weights learned through time-interval attention mechanisms. While this represents a meaningful step toward explaining flight safety events, these models are essentially ``black boxes''. The internal attention distributions, although informative, still need to be translated by the domain experts. Consequently, a substantial gap persists between model outputs and causal narratives understandable to human. This limitation constrains the practical applicability of these methods in real-world risk prevention and safety management.

The emergence of Large Language Models (LLMs) offers a promising solution to this challenge. Unlike the previous models, LLMs possess natural language reasoning capabilities, which mitigate the opacity inherent in traditional black-box machine learning methods. They can generate textual explanations aligned with their prediction outcomes~\cite{guo2024towards}, making them suitable for end-to-end interpretation of flight safety events. Nevertheless, applying LLMs to the aviation domain presents several challenges. First, QAR data consists of high-dimensional multivariate time series, whereas LLMs are primarily optimized for linguistic processing and exhibit inherent limitations in numerical reasoning~\cite{spathis2024first}. Directly feeding raw time series data into LLMs therefore leads to analytical inefficiency and unstable outputs. The cross-modal inconsistency is a large gap to be addressed. Recent studies have tried to address this issue~\cite{hu2025context,yu2023temporal,lan2025axis,guo2024towards}, which have a reference value to us. However, the complexity of QAR data is greater than that of the datasets used in these studies. 
Second, LLMs are not designed for classifying time series, so they encounter inaccuracies when dealing with classification in the domain~\cite{jin2024position}. 
Third, although fine-tuning is an effective way to improve LLMs' capability, the lack of large-scale datasets remains an obstacle, given the extremely low occurrence probability of flight safety events. 
Furthermore, due to the complexity of aviation operations and flight processes, LLMs may generate explanations that are inconsistent with established domain knowledge, leading to potential hallucination risks.

To address the aforementioned challenges, this study proposes FlightLLM, an LLM-based approach for interpretable flight safety analysis. Given that hard landing is a representative safety event during the landing stage, we choose it as an anchor in our study. The overall framework is organized around four key challenges: cross-modal inconsistency, limited classification capability, data scarcity, and insufficient domain knowledge. To deal with the first challenge, bridging the high-dimensional QAR data and the textual data, we first conduct feature engineering, compressing the raw time series into a more compact and meaningful representation. By combining automatically extracted statistical features from the TSFresh library with manually designed physical features derived from aviation domain knowledge, we build a comprehensive feature space. However, numerical features alone are still difficult for LLMs to reason. Therefore, on this basis of feature engineering, we further introduce a Semantic Discretization module, transforming numerical patterns into semantic descriptions by quantile strategy, which assists us to fully utilize the LLMs' linguistic processing strengths. Third, to compensate for the limitation of LLMs in classifying, we introduce a Statistical Expert mechanism, which utilizes the sensitivity of traditional models to numerical values to guide the LLMs' reasoning. This approach combines the precision of traditional models with the interpretability of LLMs. Furthermore, given the lack of large-scale data for fine-tuning, a Contrastive Few-Shot Learning strategy is adopted to improve this situation. We retrieve the most similar normal and hard landing samples to construct a contrastive context. By this, we compel LLMs to perform analysis between differences, which improves classification performance on vague samples without changing the internal weights of LLMs. Finally, because general-purpose LLMs lack sufficient aviation expertise, we design a structured prompt, injecting domain knowledge to mitigate the hallucination. Experimental results demonstrate that the proposed FlightLLM is not only effective in classification but also exhibits superior interpretability. Our work can help LLMs provide intuitive and high-quality textual explanations, offering valuable guidance for the analysis and prevention of incidents in civil aviation.

The main contributions of this paper are summarized as follows:

\begin{itemize}
    \item We propose FlightLLM to exploit the ability of LLMs for classification and causes analysis of flight safety events represented by hard landing. This approach can conduct a detailed causal diagnosis, while converting abstract risk assessment into practical actions, directly supporting pilot training and flight operation quality assurance. Our work bridges the gap between data-driven classification and intelligible explanation, unprecedentedly enhancing the interpretability for safety event analysis in aviation field. To the best of our knowledge, this is the first work to apply LLMs for this task.
    \item We propose a novel feature engineering that compresses complex QAR time series into refined representations by combining TSFresh-extracted statistical descriptors with manually designed physical indicators. These features are then semantized to fully leverage LLMs’ textual reasoning capabilities. These two modules address the cross-modal gap between raw time series and the natural language input required by LLMs. In addition, we introduce a traditional model as a statistical expert, whose outputs are injected as prior guidance to mitigate the inherent classification limitations of LLMs and improve prediction reliability.
    \item Experimental results on a real-world dataset consisting of 704 A320 flights demonstrate that the proposed approach exhibits not only high accuracy and precision but also unprecedented superiority in terms of interpretability. It is capable of analyzing the causes of hard landings through direct text based on persuasive evidences, which provides a reliable reference for lowering risks in future flights.
\end{itemize}

The rest of this paper is as follows: Section \uppercase\expandafter{\romannumeral2} reviews the related work. Section \uppercase\expandafter{\romannumeral3} introduces the details of FlightLLM. Section \uppercase\expandafter{\romannumeral4} shows the experiments including the setup and the analysis of the experimental results. Section \uppercase\expandafter{\romannumeral5} concludes this paper and provides potential future directions.

\section{Related Work}
\subsection{Data-Driven Flight Safety Analysis and Interpretability}
The landing phase is closely associated with flight safety. Statistics indicate that the landing phase accounts for 37\% of all fatal accidents, despite occupying only 1\% of the flight time\cite{Boeing2025Summary}. Because of this high risk probability, conducting event attribution analysis to mitigate landing-related risks remains a central focus in aviation safety management.

Local Interpretable Model-Agnostic Explanations (LIME) and Shapley Additive Explanations (SHAP) are two common methods for detecting critical features causing safety events during landing phase.
Khattak \textit{et al.}~\cite{khattak2024assessment,khattak2024estimating} proposed a TabNet framework combined with Bayesian Optimization to predict the severity of wind shear events. They respectively employed LIME and SHAP to identify critical impact factors. Midtfjord \textit{et al.}~\cite{midtfjord2022decision} utilized XGBoost to predict runway safety conditions, including temperature and pollution levels, and applied SHAP to determine influential variables. Ebensperger \textit{et al.} \cite{ebensperger2025enhancing} leveraged SHAP and feature permutation to enhance the interpretability of runway configuration assitance model.
While these studies contribute valuable insights into environmental risk assessment, their explanatory focus remains external, neglecting the internal dynamics of pilot control actions and operational procedures. However, such human factors are critical to targeted pilot training and proactive aviation safety improvement. Addressing this gap requires explaining frameworks that connect predictive results with pilot maneuvering behaviors rather than solely environmental conditions.
To advance landing safety analysis from the perspective of pilot behaviors, several studies have explored data-driven interpretability approaches. Yang \textit{et al.}~\cite{Yang2022DataDriven} adopted a hybrid feature selection strategy to filter discriminative variables and incorporated Bayesian Optimization to enhance predictive performance. They interpreted the model through SHAP-based feature importance visualization, and further analyzed the pilot actions on the basis of these figures. Similarly, Qi \textit{et al.}~\cite{qi2025hard} proposed a framework that integrates multimodal feature representations with a dual-thresholding mechanism to improve detection ability for hard landings. They utilized SHAP to interpret the model's decision logic, highlighting the importance of pilot emergency response capability.

The development of Class Activation Mapping (CAM) has attracted attention from researchers seeking to enhance interpretability in flight safety analysis. Li \textit{et al.}~\cite{li2023imtcn} proposed IMTCN, which integrates multiple Temporal Convolutional Networks (TCNs) with an improved CAM mechanism to strengthen interpretability in QAR time-series classification. IMTCN provides visual explanations for hard landing events by highlighting discriminative temporal regions through activation maps. Yang \textit{et al.}~\cite{yang2025flight} developed a flight process importance framework to assess pilot performance and employed Gradient-weighted CAM to interpret the assessment results.

In addition, Transformer architectures and attention mechanisms have become increasingly prominent in landing safety interpretability research. Shang \textit{et al.}~\cite{shang2025dual} introduced a dual-stage attention framework that provides fine-grained interpretability for hard landing prediction and attribution. Cai \textit{et al.}~\cite{cai2025fine} combined Recurrent Neural Networks (RNNs) with Transformers to model temporal dependencies and incorporated Graph Convolutional Networks (GCNs) to capture latent feature correlations, achieving high predictive accuracy. Huang \textit{et al.}~\cite{huang2025multisafe} designed a parameter importance selector based on a gating network and leveraged attention mechanisms to enhance model performance, revealing relationships between hard landing and tail strike events. 
These studies visualize attention weights to detect the parameters which significantly contribute to hard landing or tail strike. However, the interpretability still relies on post analysis by researchers. The explanation process requires domain experts to translate highlighted regions or attention distributions into meaningful causal narratives.

\subsection{Large Language Models for Time Series}
The rapid development of Large Language Models opens a new direction for time series analysis. In 2023, Yu \textit{et al.} \cite{yu2023temporal} indicated that LLMs can be applied to financial time series forecasting. Their work marks an early attempt to bridge language models and temporal data. Building on this line of inquiry, Sun \textit{et al.} \cite{sun2023test} summarized two primary strategies for utilizing LLMs in time series tasks. The first is to adapt the architecture of the LLM itself to adapt temporal inputs. The second transforms time series data into representations which pretrained large language models can directly process. Most of the subsequent studies focused on the second strategy. Hu \textit{et al.}\cite{hu2025context} proposed a Dual-Scale Context-Alignment Graph Neural Networks (DSCA-GNNs) approach. It employs GNNs to align structural and semantic information between time series and natural language representations. Chen \cite{chen2025grounding} extracted features from time series, and evaluated the performance and explanatory ability of multiple LLMs based on these features. Other works \cite{cheng2025instructime,gruver2023large,jin2023time} attempted to unify time series and linguistic modalities through vector quantization and text embeddings, which map numerical sequences into token-like representations. In contrast, some studies \cite{lee2025timecap,guo2024towards} directly converted time series into textual descriptions and constructed prompts aimed at specific tasks to forecast and interpret. Beyond forecasting, anomaly detection for time series is also a major research direction. Tian \textit{et al.} \cite{lan2025axis} introduced an AXIS approach, which uses Transformers to extract embeddings from time series and projects them into the semantic space of LLMs to detect abnormal segments. Liu \textit{et al.} \cite{liu2025large} combined Few-Shot Learning with an Anomaly Detection Chain-of-Thought (AnoCoT) strategy to activate the reasoning and explanation capabilities of LLMs. This method reports strong empirical performance. To provide an evaluation perspective systematically, Li \textit{et al.}\cite{li2025stbench} examined LLMs' performance across four dimensions, i.e., knowledge comprehension, spatio-temporal reasoning, accurate computation, and downstream applications. The four aspects offer a structured assessment for LLMs' ability to process spatio-temporal data.
Although the datasets used in these studies are less complex than QAR data, which typically contain higher dimensionality and a larger number of variables, these studies still provide valuable and innovative perspectives for the analysis of QAR time-series data.

\section{Methodology}
\begin{figure*}[htbp]
\centering
\includegraphics[scale=0.14]{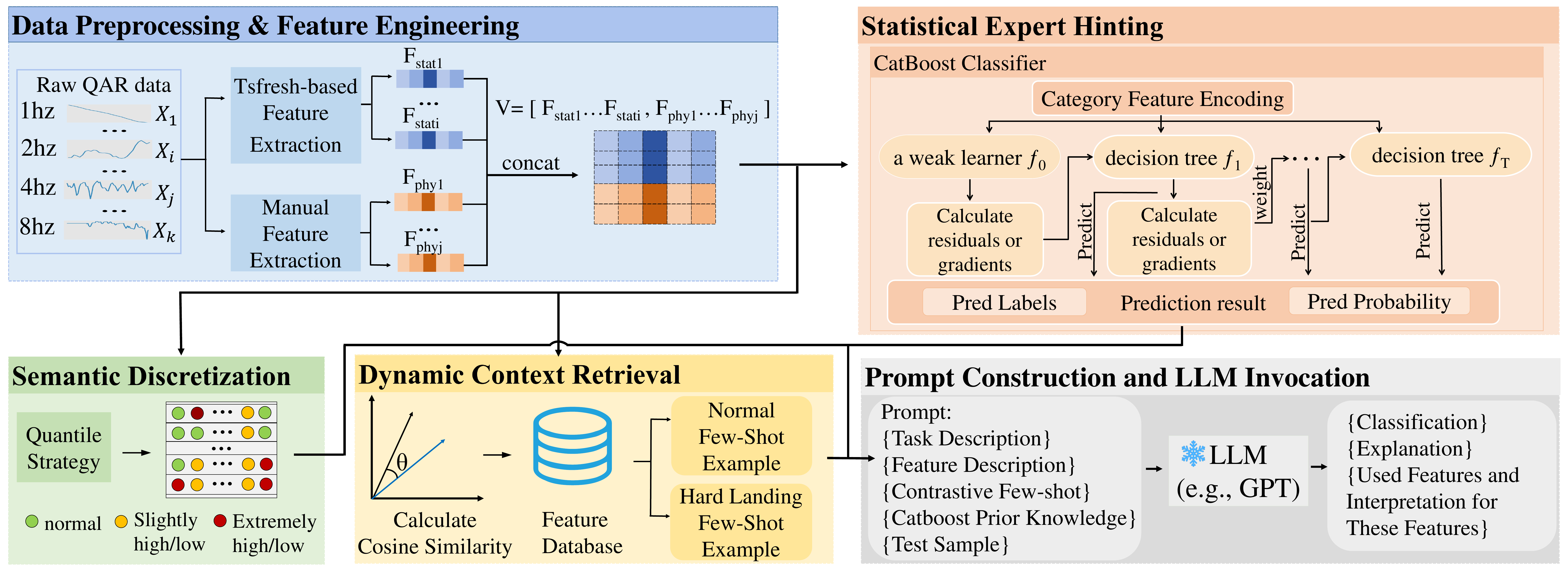}
\caption{The framework of FlightLLM consists of five modules: Data Preprocessing and Feature Engineering, Semantic Discretization, Statistical Expert Hinting, Dynamic Context Retrieval and Prompt Construction and LLM Invocation. Statistical Expert denotes CatBoost model.}
\label{framework}
\end{figure*}

We first introduce the overall architecture of the proposed FlightLLM, as shown in Fig. \ref{framework}. The framework consists of five functional modules: (\romannumeral1) Data Preprocessing and Feature Engineering; (\romannumeral2) Semantic Discretization; (\romannumeral3) Statistical Expert Hinting; (\romannumeral4) Dynamic Context Retrieval;  (\romannumeral5) Prompt Construction and LLM Invocation. Among these components, Modules~(\romannumeral1)–(\romannumeral4) form the core pipeline of the system, while Module~(\romannumeral5) serves as the interface between structured aviation data and the large language model. The detailed design and implementation of these modules will be presented in the subsequent sections.

\subsection{Data Preprocessing and Feature Engineering}
The objective of hard landing classification is to determine, based on QAR data, whether the vertical acceleration (VRTG) will exceed a predefined safety threshold. The prediction relies on multiple flight parameters, including \textit{PITCH}, \textit{ALT\_STD}, and \textit{TLA}, among others. If the predicted VRTG exceeds the threshold, the sample is labeled as hard landing. Otherwise, it is classified as a normal landing. It should be noted that the threshold for hard landing identification varies across aircraft types. In this study, we use real-world flight data collected from A320 aircraft. Following prior studies\cite{li2023imtcn,cai2025fine,huang2025multisafe}, we set the threshold $\theta$ to $1.5g$. A hard landing event is defined as occurring when Equation \ref{vrtg} is satisfied:

\begin{equation}
\label{vrtg}
\text{VRTG}_{[altitude \leq 0]} \geq \theta
\end{equation}

QAR data constitute high-dimensional multivariate time series with different sampling frequencies and substantial data volume. Such characteristics make them difficult to process directly using LLMs. Therefore, preprocessing is required to bridge the gap between structured flight data and language-based reasoning models. After standard data cleaning, we define the observation window as the interval from 30 seconds before touchdown to the touchdown moment. The dataset is denoted as ${D} = \{(\mathbf{X}_i, \mathbf{Y}_i)\}_{i=1}^{N}$, where $N$ is the total number of flight samples. Each sample $\mathbf{X}_i \in \mathbb{R}^{L \times C}$ represents a multivariate time series matrix, with $L$ denoting the sequence length and $C$ denoting the number of sensor channels. At time step $t \in [1, L]$, the observation vector is $\mathbf{x}_t \in \mathbb{R}^C$. The label $Y_i \in \{0,1\}$ indicates whether a hard landing event occurs.

To construct a discriminative and interpretable feature space, we adopt a dual-branch extraction strategy. To capture statistical features with high importance, we employ the \textit{TSFresh} library\cite{christ2018time} for automated feature extraction. Let $\mathcal{T}(\cdot)$ denote the extraction operator. For each sample $\mathbf{X}_i$, a candidate statistical feature set is generated. To mitigate complexity and control dimensionality, we group features according to their associated sensors and conduct statistical significance tests. The top-$k$ significant features are retained as:
\begin{equation}
\label{tsfresh}
\mathbf{F}_{stat} = \text{Select}_k\big(\mathcal{T}(\mathbf{X}_i)\big) \in \mathbb{R}^{d_{stat}}.
\end{equation}

Although $\mathbf{F}_{stat}$ includes statistical features with high importance scores, many of these features are abstract. They lack intelligible physical meaning, which makes them difficult to be comprehended by humans and LLMs. Therefore, this abstraction limits the LLMs' ability to reason. To enhance semantic clarity, we introduce a feature branch based on aviation domain knowledge. We explicitly extract indicators, such as minimum descent rate from 20 feet to touchdown, to form a physically interpretable feature vector $\mathbf{F}_{phy} \in \mathbb{R}^{d_{phy}}$.

Finally, we integrate statistical robustness and physical interpretability through features concatenation. The resulting hybrid feature vector is defined as:
\begin{equation}
\mathbf{V} = \mathbf{F}_{stat} \oplus \mathbf{F}_{phy} 
\in \mathbb{R}^{d_{stat} + d_{phy}}.
\tag{2}
\end{equation}

The detailed composition of the constructed feature space is summarized in Table \ref{features}.

\begin{table*}[!htbp]
\caption{Detailed Description of the Extracted Hybrid Feature Space}
\label{features}
\centering
\begin{tabularx}{\textwidth}{l l X}
\hline
\textbf{Category} & \textbf{Feature Symbol} & \textbf{Description} \\ \hline

\multicolumn{3}{c}{\textbf{Part A: Automated Statistical Features ($\mathbf{F}_{stat}$)}} \\ \hline

\multirow{4}{*}{\textbf{Vertical Speed}} 
& $Var(\Delta \text{IVV}_{0.4}^{1.0})$ & The variance of vertical speed in the 40th to 100th percentile range. \\
& $Var(|\Delta \text{IVV}|)$ & The absolute variance of the vertical speed change within the full percentile range. \\
& $Var(\Delta \text{IVV}_{max-min})$ & The variance between the maximum and minimum changes in vertical speed. \\
& $\mathcal{F}_{real}^{12}(\text{IVV})$ & The real frequency energy of the vertical speed in the higher frequency band. \\ \hline

\multirow{8}{*}{\textbf{Altitude}} 
& $Q_{0.1}(\text{RADIO}_{LH})$ & The 10th percentile value of the left radio height. \\
& $\bar{\Delta}_{abs}(\text{RADIO}_{RH}^{0.0-0.2})$ & The average absolute change of the right radio height at the lowest percentile. \\
& $r_{min}^{L=50}(\text{ALT}_{STD})$ & The minimum correlation of standard altitude in the linear trend of the long window. \\
& $\mathcal{F}_{imag}^{4}(\text{ALT}_{STD})$ & The imaginary frequency energy of the standard altitude in the low and medium frequency bands. \\
& $\nabla_{min}^{L=50}(\text{ALT}_{QNH})$ & The minimum slope in the long-term trend of altitude. \\
& $\beta_{min}^{L=50}(\text{ALT}_{QNH})$ & The minimum intercept in the long-term trend of altitude. \\
& $\bar{\Delta}_{abs}(ALT_{QNH}^{0.0-0.2})$ & The mean absolute change of QNH altitude within the lowest 20th quantile. \\
& $Count(ALT_{QNH} > 0)$ & The total count of valid altitude data points within the entire flight phase. \\ \hline

\multirow{5}{*}{\textbf{Pitch}} 
& $CV(\text{PITCH}_{cmd})$ & Pitch command variation coefficient. \\
& $Var(\Delta \text{PITCH}_{0.8}^{1.0})$ & The variance of the pitch change within the 80th to 100th percentile range. \\
& $Var(\nabla^{L=5}(\text{PITCH}))$ & The variance of the slope of the short segment linear trend of the pitch. \\
& $\rho_{\tau=4}(\text{PITCH})$ & The autocorrelation of the pitch with a 4-step delay. \\
& $\text{Loc}_{max}(\text{PITCH})$ & The position where the maximum value of the pitch first occurs. \\ \hline

\multirow{2}{*}{\textbf{Throttle}} 
& $\mathcal{F}_{imag}^{42}(\text{TLA}_2)$ & The energy of the throttle 2 in the mid-high frequency range. \\
& $\bar{\nabla}^{L=50}(\text{TLA}_1)$ & Mean of the long-term slope of the throttle 1. \\ \hline

\multicolumn{3}{c}{\textbf{Part B: Domain-Driven Physical Features ($\mathbf{F}_{phy}$)}} \\ \hline

\multirow{2}{*}{\textbf{Descent Timing}} 
& $\Delta t_{50\to TD}$ & The time from 50 feet to touchdown. \\
& $\Delta t_{20\to TD}$ & The time from 20 feet to touchdown. \\ \hline

\multirow{3}{*}{\textbf{Kinematics (IVV)}} 
& $\text{IVV}_{min}^{50\to TD}$ & The minimum vertical speed from 50 feet to touchdown. \\
& $\text{IVV}_{min}^{20\to TD}$ & The minimum vertical speed from 20 feet to touchdown. \\
& $\overline{\text{IVV}}^{50\to TD}$ & The mean vertical speed from 50 feet to touchdown. \\ \hline

\multirow{7}{*}{\textbf{Attitude (Pitch)}} 
& $\text{PITCH}_{max}^{50\to TD}$ & The maximum pitch from 50 feet to touchdown. \\
& $\text{PITCH}_{min}^{50\to TD}$ & The minimum pitch from 50 feet to touchdown. \\
& $\max|\dot{\text{PITCH}}^{20\to TD}|$ & The maximum rate of pitch change from 20 feet to touchdown. \\
& $\text{PITCH}_{TD}$ & The pitch at touchdown. \\
& $t(\text{PITCH}_{max})$ & The moment when the maximum pitch occurs. \\
& $\Delta \text{PITCH}_{cmd}^{20\to TD}$ & The variation amount of the pitch command from 20 feet to touchdown. \\
& $\max|\dot{\text{PITCH}}_{cmd}^{20\to TD}|$ & Maximum rate of change of pitch command from 20 feet to touchdown. \\ \hline

\multirow{5}{*}{\textbf{Airspeed \& Energy}} 
& $\text{IAS}_{50ft}$ & The indicated airspeed at 50 feet. \\
& $\text{IAS}_{TD}$ & The indicated airspeed at touchdown. \\
& $\Delta V_{50ft}$ & Speed deviation at 50 feet. \\
& $\Delta V_{TD}$ & Speed deviation at touchdown. \\
& $\overline{E_k}^{20\to 5}$ & Average energy in the final stage (20 to 5 feet). \\ \hline

\multirow{4}{*}{\textbf{Lateral / Control}} 
& $\max|\text{ROLL}^{50\to TD}|$ & The maximum absolute value of the roll angle from 50 feet to touchdown. \\
& $|\text{ROLL}_{TD}|$ & The absolute value of the roll angle at touchdown. \\
& $\max|\text{RUDD}^{50\to TD}|$ & The maximum absolute value of rudder position from 50 feet to touchdown. \\
& $\overline{\dot{\text{TLA}}}_{1s}$ & The rate of throttle change one second before touchdown. \\ \hline

\multirow{3}{*}{\textbf{Environmental}} 
& $\max|\text{WIND}_{lat}|$ & The maximum lateral wind component from 50 feet to touchdown. \\
& $\text{WIND}_{lon}^{TD}$ & The longitudinal wind component at touchdown. \\
& $\max(\text{WIND}_{spd})$ & The maximum speed of the wind from 50 feet to touchdown. \\ \hline
\end{tabularx}
\end{table*}

\subsection{Semantic Discretization}
The tokenizer architecture of LLMs is designed for natural language rather than continuous numerical values. Consequently, their ability to process raw numerical data is generally weaker than their capability to analyze text\cite{spathis2024first}. In practice, numbers with multiple digits are often sliced into individual sub-tokens. The token fragmentation destroys numerical continuity and weakens the internal physical meaning, resulting in semantic incoherence. Additionally, the lack of contextual information makes it difficult for LLMs to interpret numerical features accurately. For example, giving an isolated input such as ``IVV = $-$800 ft/min'', an LLM may struggle to assess its severity without additional contextual cues. The model can hardly recognize whether this value indicates a normal descent rate or a potentially dangerous condition. This limitation highlights the gap between number and semantics.

To bridge the gap between numerical domain and semantic domain, we propose a Semantic Discretization module, utilizing a semantic mapping strategy based on quantile.  We transform continuous flight features into qualitative expressions derived from their statistical distribution. Let $\mathcal{X}$ denote the global distribution of a specific feature, then We define a set of quantization thresholds $\boldsymbol{\tau} = [\tau_1, \tau_2, \tau_3, \tau_4]$, representing the $5^{th}, 25^{th}, 75^{th},$ and $95^{th}$ percentiles of $\mathcal{X}$, respectively. Furthermore, $\mathcal{S} = \{s_1, s_2, s_3, s_4, s_5\}$ represents the ordered set of semantic tokens: \textit{Extremely Low, Slightly Low, Normal, Slightly High,} and \textit{Extremely High}. The semantic discretization function $\Phi: \mathbb{R} \to \mathcal{S}$ maps an input feature $f$ to its corresponding semantic token as follows:

\begin{equation}
\Phi(f) = 
\begin{cases} 
s_1, & f \le \tau_1 \\
s_2, & \tau_1 < f \le \tau_2 \\
s_3, & \tau_2 < f < \tau_3 \\
s_4, & \tau_3 \le f < \tau_4 \\
s_5, & f \ge \tau_4
\end{cases}
\label{eq:semantic_mapping}
\end{equation}

Based on the mapping strategy, we construct a structured rich semantic descriptor for each feature. This descriptor includes three components: (i) the physical meaning of the feature; (ii) the semantic label based on the quantile; (iii) its raw numerical value. 
This hybrid expression with qualitative and quantitative details provides two key advantages. First, before the LLM processes the raw numerical value, the semantic label can activate its prior knowledge, which improves its reasoning ability. The model can therefore form an initial assessment of the flight condition at a conceptual level, rather than relying solely on interpreting numbers. Second, by explicitly encoding the degree of statistical deviation, the expression removes the need for the LLM to perform implicit numerical comparisons. The abnormal level is predefined through distributional statistics, which reduces reliance on the model’s limited arithmetic precision. Consequently, the LLM can focus on attribution reasoning and analysis.

\subsection{Statistical Expert Hinting}
Leveraging a small model to guide large models is a weak-to-strong learning method to improve the stability and reliability of LLM outputs\cite{burns2023weak}. In this study, we introduce CatBoost as an auxiliary statistical expert to provide structured prior guidance. CatBoost is a gradient boosting model whose base learners are decision trees. During training, the model computes gradients according to Equations as follows:

\begin{equation}
h^t = \underset{h \in H}{\arg \min} \, \mathcal{L}(F^{t-1} + h) = \underset{h \in H}{\arg \min} \, \mathbb{E} L(y, F^{t-1}(\mathbf{x}) + h(\mathbf{x})).
\end{equation}

\begin{equation}
h(\mathbf{x}) = \sum_{j=1}^{J} b_j \mathbbm{1}_{\{\mathbf{x} \in R_j\}},
\end{equation}
 where $b_j$ denotes the predicted value of the $j$-th leaf node and $R_j$ represents the sample region associated with that leaf. Based on the calculated gradients, CatBoost constructs a new balanced decision tree to fit the residual errors. Then the model is updated according to the equation as follows:

\begin{equation}
    F^t = F^{t-1} + \alpha h^t
\end{equation}
Unlike traditional Gradient Boosting Decision Trees (GBDT), CatBoost does not rely on one-hot encoding for categorical variables. Instead, it transforms categorical features into numerical representations using the encoding formula defined as follows:

\begin{equation}
    X_{\sigma_{p,k}} = \frac{\sum_{j=1}^{p-1} [X_{\sigma_{j,k}} = X_{\sigma_{p,k}}] Y_{\sigma_s} + \beta \cdot P}{\sum_{j=1}^{p-1} [X_{\sigma_{j,k}} = X_{\sigma_{p,k}}] + \beta}
\end{equation}
where $P$ denotes the prior value and $\beta$ denotes the weight assigned to the prior. This encoding strategy reduces information loss, thereby improving boosting efficiency. In addition, CatBoost also combines features as a new one. After the first split of a tree, it will use a greedy method to consider all splits selected in the tree as a category. This design enables the model to capture more complex dependencies. Empirical studies have demonstrated that CatBoost often performs better than XGBoost and LightGBM in multiple classical machine learning tasks\cite{Prokhorenkova2018CatBoost}. It can well suit for providing statistical guidance within our study because of its robustness and specific feature importance estimation.

In our approach, the extracted features are input to CatBoost and the predictions generated by CatBoost serve as statistical anchors, providing reliable reference points that elevate the reasoning capability of LLMs. This design follows a weak-to-strong conception rather than replacing the LLM’s decision process. We inject the predicted label and the associated probability produced by CatBoost into the prompt as auxiliary contextual information. The LLM does not just blindly follow the prediction given by CatBoost, but can take advantage of the signal. When the CatBoost prediction is consistent with the LLM’s judgment, the model can incorporate this agreement to strengthen the credibility of its explanation. When inconsistencies arise, the LLM is encouraged to examine the evidences again. This reflection helps the LLM provide a reasoned justification, either by correcting its own inference or by explicitly challenging the CatBoost's output. Overall, the collaboration enhances predictive ability and improves reliability. More importantly, it demonstrates how LLMs can extend traditional models, which is a referable direction for analysis tasks of flight safety events.

\subsection{Dynamic Context Retrieval}
Traditional time-series classification approaches generally rely on large-scale labeled datasets for supervised training. These methods not only involve high training costs but also face difficulties in adapting to new scenarios. In the domain of flight safety, high-risk events such as hard landings occur with extremely low frequency, making it impractical to collect sufficient data for large-scale fine-tuning.
Under such conditions, the strong In-Context Learning (ICL) capability of LLMs becomes particularly valuable, as it allows them to handle specialized tasks with only a few representative examples and infer the boundary between normal and abnormal samples~\cite{liu2025large}. Therefore, even in the absence of massive datasets, competitive performance still can be achieved. Based on ICL, we propose a Dynamic Context Retrieval strategy, which dynamically selects representative positive and negative samples as contrastive references.
We first construct a reference database 
\[
D=\{(x_j, y_j)\}_{j=1}^{N},
\]
where $x_j$ denotes the feature vector of a historical flight sample and $y_j$ denotes its corresponding landing status label. 

\begin{figure*}
\centering
\subfloat[]{\includegraphics[width=2.5in]{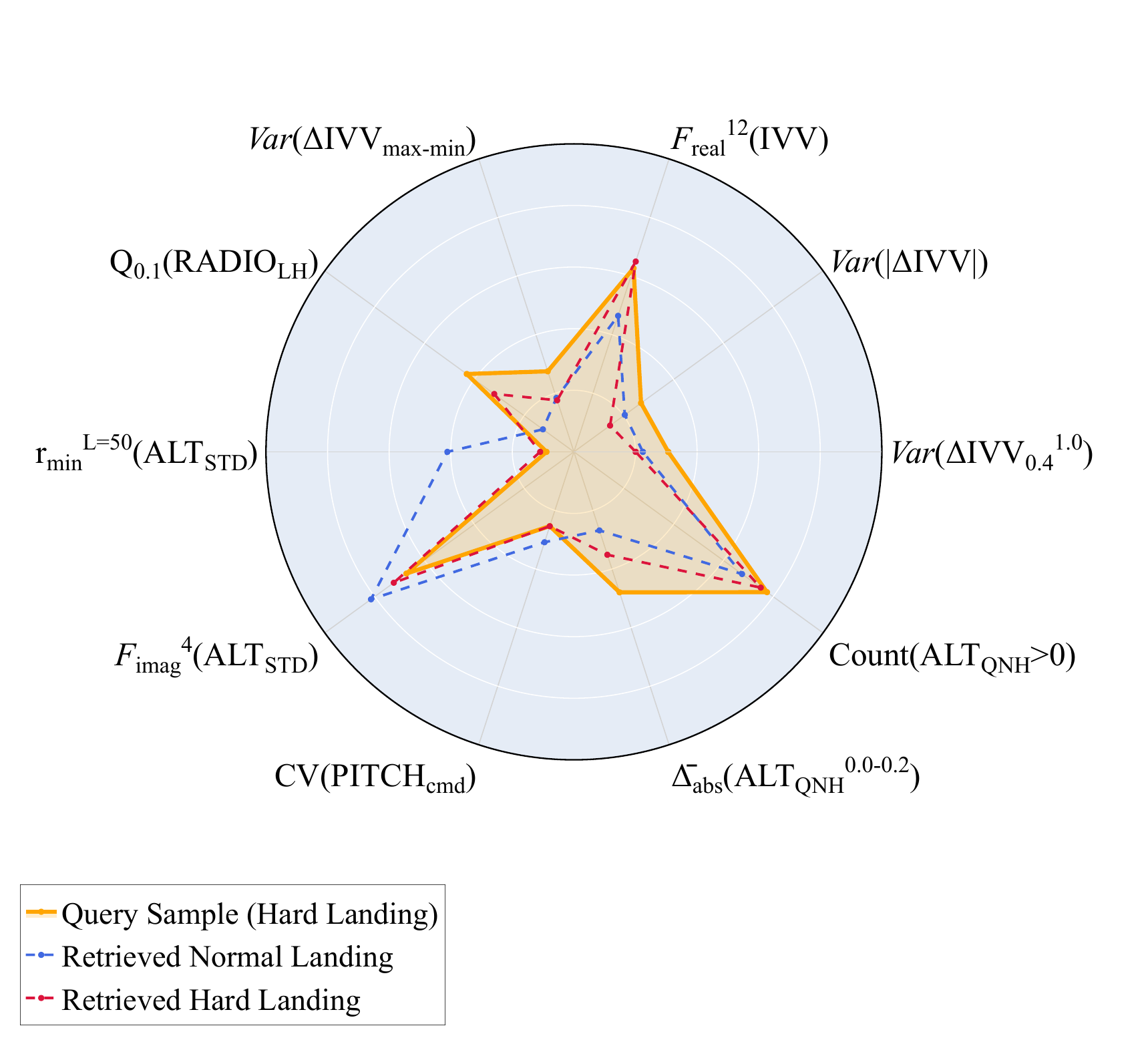}}
\label{radar639}
\hfil
\subfloat[]{\includegraphics[width=2.5in]{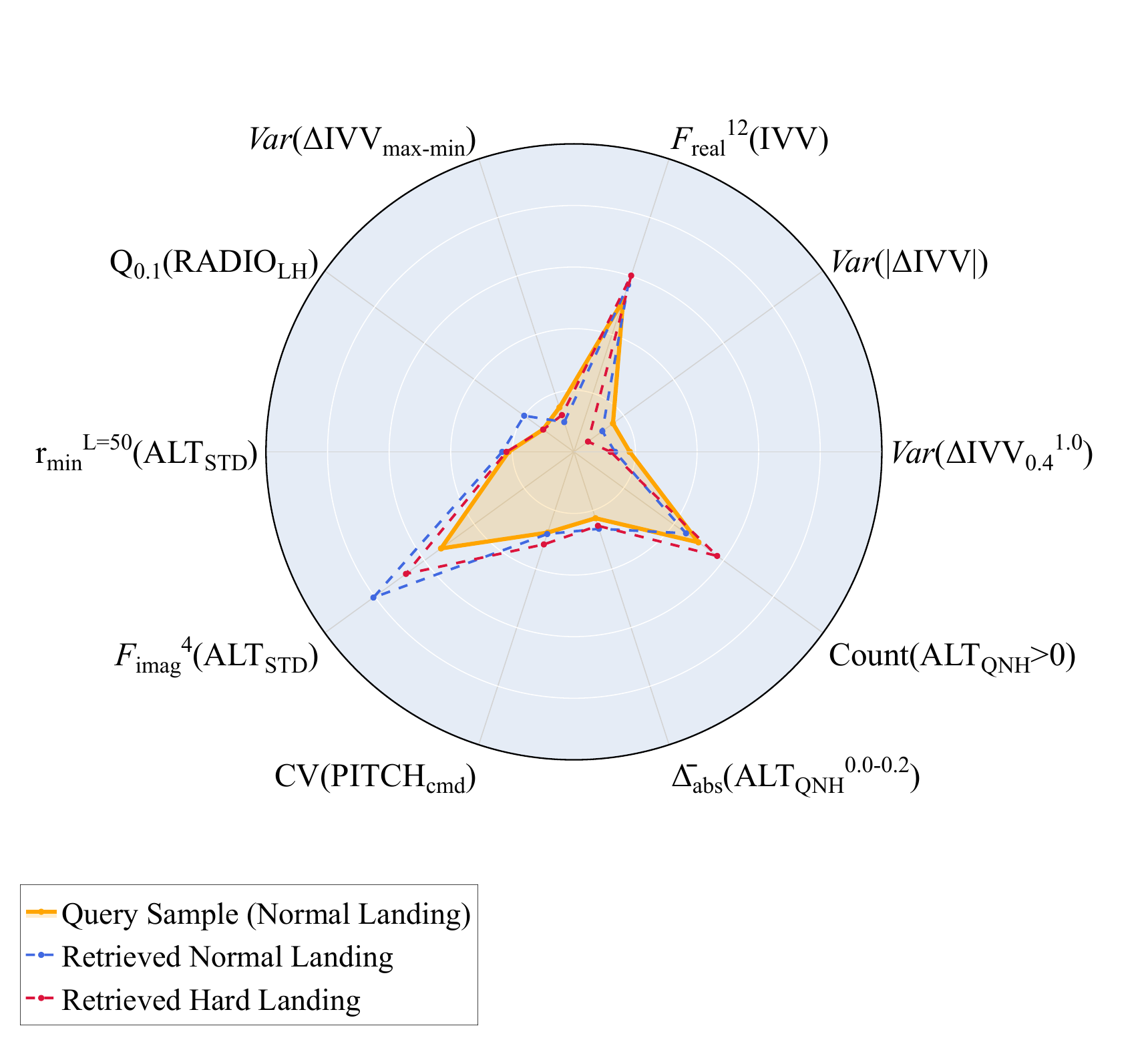}}
\label{radar277}
\caption{High-Dimensional Feature Profile Match}
\label{radar}
\end{figure*}

During the process of constructing prompt, for a query sample $x_{query}$, we dynamically retrieve similar instances from $D$ based on similarity in vector space. Cosine Similarity is adopted as the similarity metric. The similarity score $CS$ between $x_{query}$ and a candidate sample $d_q \in D$ is computed according to Equation as follows:

\begin{equation}
CS(x_{query},d_q)= \frac{{\textstyle \sum_{i=1}^{n}}x_{query_i}d_{q_i}}{\sqrt{ {\textstyle \sum_{i=1}^{n}x_{query_i}^2}}\sqrt{ {\textstyle \sum_{i=1}^{n}d_{q_i}^2} }   } 
\end{equation}
where $x_{query_i}$ and $d_{q_i}$ represent the $i$-th components of the respective feature vectors. Based on the similarity score, we select two samples as references: the most similar normal landing instance $x_{norm}$ and the most similar hard landing instance $x_{hard}$. The equations we select samples are as follows:
\begin{align}
\mathbf{x}_{norm} &= \mathop{\text{argmax}}_{\mathbf{x}_j \in \mathcal{D}_{normal}} CS(\mathbf{x}_{query}, \mathbf{x}_j) \\
\mathbf{x}_{hard} &= \mathop{\text{argmax}}_{\mathbf{x}_k \in \mathcal{D}_{hard}} CS(\mathbf{x}_{query}, \mathbf{x}_k)
\end{align}
To visualize the similarity between the retrieved samples and query samples, a hard landing query sample and a normal landing query sample are shown in Fig. \ref{radar}(a) and Fig. \ref{radar}(b) as instances. We only draw the top ten features in the figures for clarity. 
After retrieval, the selected samples are injected into the prompt as Few-Shot examples, forming a contrastive reasoning context. Because the retrieved samples are highly similar to $x_{query}$, the LLM can identify key factors by comparing subtle differences, clarifying the decision boundary within the reasoning process.

\subsection{Prompt Construction and LLM Invocation}
A major challenge in applying LLMs to flight safety analysis lies in their insufficient specialized knowledge of aviation dynamics, flight mechanics, and pilot control logic. LLMs may misunderstand the semantic meanings of the extracted features and their relationships with aircraft maneuvers. Such semantic gaps can lead to unstable reasoning and wrong explanations. To address this issue, we perform feature description engineering. Specifically, we provide explicit textual explanations for the extracted features. Each feature is linked to corresponding pilot control behaviors or flight dynamics. This rich semantic structure enables LLMs to associate numerical descriptors with meaningful contexts. By embedding knowledge into the prompts, we improve the reliability and interpretability of the generated outputs. This design reduces ambiguity between features and options, assisting the LLM keep reasoning consistent with aviation field.

\begin{figure}[htbp]
\centering
\includegraphics[width=3.0in]{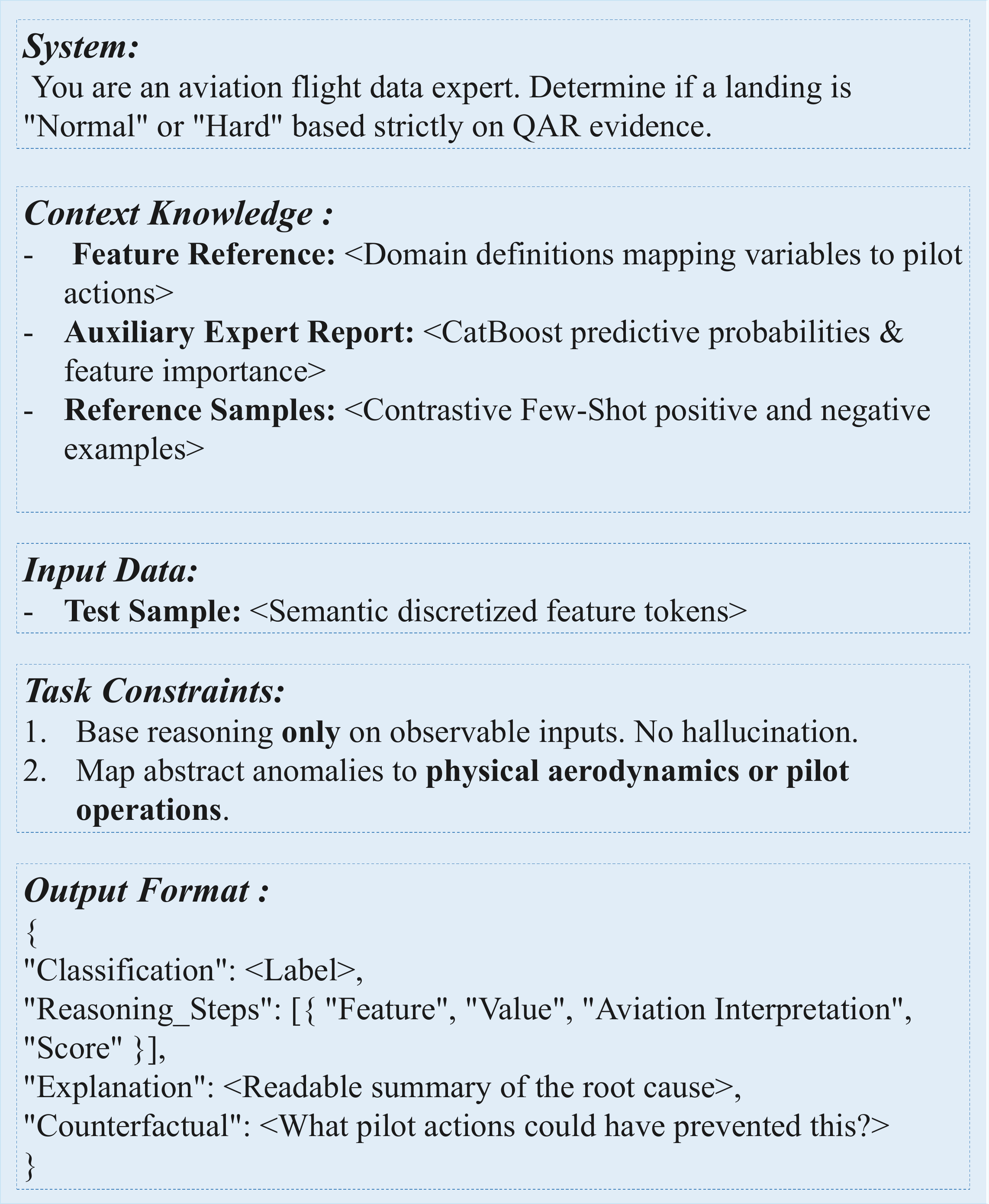}
\caption{The Design of the Prior-Guided Prompt Template.}
\label{prompt}
\end{figure}

In addition, Chain-of-Thought (CoT) prompting is a technique that explicitly guides LLMs to generate intermediate reasoning steps, thereby enhancing their analytical performance \cite{wei2022Chain}. In this study, we adopt CoT to guide LLMs to analyze flight parameters step by step, interpret their operational significance, evaluate their consistency with hard landing mechanisms, and then derive a final judgment.
This structured reasoning paradigm not only compensates for LLMs' lack of inherent aviation expertise but also aligns the inference process with the analytical workflow of flight safety specialists. 
The overall structure of the designed prompt is shown in Fig. \ref{prompt}.

\begin{table*}[!htbp] 
\caption{The Design of the Prior-Guided Prompt Template}
\label{tab:prompt_design}
\centering
\begin{tabular}{@{}lp{13cm}@{}}
\toprule
\textbf{Prompt Module} & \textbf{Content / Instruction Summary} \\ \midrule

\textbf{1. System Persona} 
& \textit{``You are an expert in aviation flight data. Your task is to determine whether a landing is normal or hard based on strict evidence.''} \\ \addlinespace

\textbf{2. Context Injection} 
& \textbf{Feature Dictionary:} Inject semantic definitions mapping abstract statistical variables to specific aircraft states. \\
& \textbf{Auxiliary Report ($\mathcal{H}$):} Inject the statistical prior distribution generated by the CatBoost expert model. \\
& \textbf{Few-Shot Examples ($\mathcal{E}$):} Provide historical normal and hard landing prototypes to construct a contrastive learning context. \\ \addlinespace

\textbf{3. Input Formulation} 
& Test Sample Features ($\mathbf{V}_{test}$): The sequence of semantic tokens representing the current flight parameters. \\ \addlinespace

\textbf{4. Task Constraints} 
& 1. Rely \textbf{strictly} on provided observable features; avoid speculation or hallucination. \\
& 2. Relate statistical feature anomalies directly to \textbf{pilot operations} and aerodynamics. \\
& 3. Explicitly state ambiguity if feature evidence is contradictory. \\ \addlinespace

\textbf{5. Output Schema} 
& Enforce a strict JSON output encompassing three cognitive steps: \\
& - \texttt{Classification}: Predicted label (Normal / Hard). \\
& - \texttt{Reasoning\_Chain}: A step-by-step physical attribution for each abnormal feature, alongside its importance score. \\
& - \texttt{Counterfactual}: Conditions under which the current landing classification would change, providing actionable flight safety insights. \\ \bottomrule
\end{tabular}
\end{table*}

\section{Experiments}
In this section, we conduct comprehensive experiments to evaluate the effectiveness of the proposed FlightLLM. We select hard landing as the anchor event for experiments. As a representative high-risk flight safety event, hard landing involves complex causal mechanisms. Its complexity and strongly coupled characteristics make it an ideal benchmark to rigorously evaluate both the classification capability and the attribution quality of FlightLLM. Demonstrating effectiveness on such a challenging scenario provides a solid foundation for extending flightLLM to other flight safety events, such as tail strikes and runway excursions, which share similar operational and causal complexities.

To avoid data leakage, we define a dynamic time point $t$ as the moment of complete touchdown, namely, when all landing gears touch ground. For each flight, the data sequence is sliced to the interval from 30 seconds before touchdown to the dynamic time point $t$. This segmenting strategy prevents signals after touchdown from influencing the cause analysis, preserving causal consistency in the evaluation process.    

 \subsection{Dataset and Experimental Configuration}
According to the criterion defined in Equation\ref{vrtg}, we identified 282 hard landing samples from 37,929 real-world A320 flight records. A relatively balanced dataset is more reasonable, so we randomly selected 422 normal landing samples. The two parts constitute the final experimental dataset of 704 flights, with a hard-to-normal ratio of approximately 2:3. This sampling strategy is motivated by the objective of the study. Our primary focus is the attribution analysis of hard landing events rather than large-scale normal pattern modeling. Applying the LLM to all 37,647 normal landing samples is unnecessary because it would result in substantial computational overhead meanwhile providing disproportionate analytical value. Therefore, we retain a representative subset of normal flights to support contrastive reasoning while maintaining computational feasibility. 

For each flight, the full trajectory from engine start to shutdown is stored in an individual CSV file. Each file contains 32 flight parameters with different sampling frequencies, as summarized in Table \ref{params}. To unify inconsistent frequencies of different parameters from multiple channels, we perform a resampling procedure that standardizes all parameters to the same sampling rate of 4Hz.
\begin{table*}[]
\centering
\caption{The parameters from QAR data}
\label{params}
\begin{tabular}{llcllc}
\hline
Parameter  & Description                  & Frequency(Hz) & Parameter & Description                 & Frequency(Hz) \\ \hline
ALT\_QNH   & Altitude                     & 1             & ROLL      & Roll angel                  & 2             \\
ALT\_STD   & Standard altitude corrected  & 1             & ROLL\_CPT & ROLL\_CPT                   & 8             \\
RADIO\_LH  & Left radio height            & 4             & ROLL\_FO  & Deputy captain roll control & 8             \\
RADIO\_RH  & Right radio height           & 4             & HEAD\_MAG & Magnetic heading direction  & 1             \\
LDGL       & Left landing gear state      & 4             & WIN\_DIR  & Wind direction              & 1             \\
LDGR       & Right landing gear state     & 4             & WIN\_SPD  & Wind speed                  & 1             \\
LDGNOS     & Nose landing gear state      & 4             & RUDD      & Rudder position             & 2             \\
IAS        & Indicated airspeed           & 1             & N11       & Engine 1 speed ratio        & 1             \\
VAPP       & Landing reference speed      & 1             & N12       & Engine 2 speed ratio        & 1             \\
GS         & Ground speed                 & 1             & TLA1      & Throttle lever 1 position   & 1             \\
VRTG       & Vertical acceleration        & 8             & TLA2      & Throttle lever 2 position   & 1             \\
IVV        & Vertical speed               & 1             & FLAP\_PL  & Left flap actual angle      & 1             \\
PITCH      & Pitch angle                  & 4             & FLAP\_PR  & Right flap actual angle     & 1             \\
PITCH\_CPT & Captain pitch control        & 8             & DME1      & DME 1 distance              & 1             \\
PITCH\_FO  & Deputy captain pitch control & 8             & DME2      & DME 2 distance              & 1             \\
GW         & Aircraft gross weight        & 1             & VAPP      & Speed reference             & 1             \\ \hline
\end{tabular}
\end{table*}

The large language models selected for this experiment include GPT-3.5, DeepSeek-V1, and GLM-4.7-flash. The performance of all models is evaluated using the following metrics: Accuracy, Precision, Recall, and F1-Score.

\subsection{Baselines}
The baseline models compared with FlightLLM in our experiments are as follows.

\begin{itemize}
    \item \textbf{LSTM}\cite{tong2018innovative}: All QAR parameters are first downsampled to 1Hz. A Long Short-Term Memory (LSTM) network is then employed for classification. LSTM introduces memory blocks, which can capture long-term temporal dependencies. Each memory block includes memory cells, a set of multiplicative gates and the peephole connections, enabling effective modeling of sequential flight data.
    \item \textbf{SVM}\cite{Hu2016SVM}: A classifier based on SVM is applied to process flight data segments sliced based on height. Recursive Feature Elimination (RFE) is adopted for feature selection, while a grid-search strategy is used to optimize hyperparameters.
    \item \textbf{RF}\cite{sun2021risk}: The data is first sliced based on height and then balanced using the SMOTE. A Random Forest classifier, composed of decision trees, is then trained to perform hard landing classification.
    \item \textbf{KNN}\cite{lee2012nearest}: The time series are segmented into fixed-length windows. The KNN algorithm computes similarity between windows and assigns the class label based on the majority vote among the $k$ nearest neighbors.
    \item \textbf{CNN}\cite{zhao2017convolutional}: A CNN is used to automatically learn temporal features through convolution and pooling operations. The features are subsequently fed into a multilayer perceptron (MLP) for classification.
    \item \textbf{IMTCN}\cite{li2023imtcn}: IMTCN is an interpretable model based on multiple Temporal Convolutional Networks (TCNs). It integrates Class Activation Mapping (CAM) to enhance transparency by identifying variables that contribute most to classification decisions.
    \item \textbf{SDTAN}\cite{chen2023sdtan}: SDTAN is built upon a time-interval attention mechanism. The model consists of a STG encoder, a variable correlation extraction block and a variable selection module to capture dynamic relationships between flight parameters.
    
\end{itemize}

\section{Results and Discussion}
\subsection{Classification Performance}
We evaluate the proposed FlightLLM using three backbone models: GLM, GPT, and DeepSeek. Table \ref{class} compares our approach with seven baseline methods. Bold values indicate the best performance for each metric. According to Table \ref{class}, both FlightLLM-GPT and FlightLLM-Deepseek achieve an Accuracy of 81.56. This value exceeds that of traditional machine learning models such as SVM and RF and more complex deep learning architectures such as IMTCN and SDTAN. These results indicate that FlightLLM can handle the hard landing classification task effectively.
Notably, FlightLLM-Deepseek achieves the highest Precision among all evaluated models, reflecting the model’s ability to suppress False Positives. A high Precision implies that predicted hard landing events are highly credible, which is critical in operational risk management. 

Both CNN and SDTAN demonstrate strong performance. CNN achieves the highest recall, while SDTAN attains the highest F1 score. This performance can be attributed to the convolutional operations.
Convolution kernels slide across the input data and share weights, which means the model needs to learn far fewer parameters. With fewer parameters to optimize, the model is less likely to memorize the training data and can generalize more effectively on small datasets.
In contrast, IMTCN equips multiple TCN channels for QAR parameters with different frequencies, which causes the sharp increase in the number of parameters within the model. Consequently, it requires large-scale datasets for stable parameter optimization. Given the relatively small dataset size (704 samples) in this study, its capability on the test set is constrained, which explains its comparatively weaker performance.

\begin{table}[]
\caption{Classification results of different methods}
\label{class}
\begin{tabular}{ccccc}
\hline
Model            & Accuracy       & Precision      & Recall         & F1             \\ \hline
LSTM             & 56.03          & 45.95          & 60.71          & 52.31          \\
SVM              & 57.72          & 46.82          & 54.87          & 50.53         \\
RF               & 56.74          & 50.88          & 46.77          & 48.74          \\
KNN              & 54.61          & 34.62          & 16.07          & 21.95          \\
CNN              & 77.30          & 67.14          & \textbf{83.93} & 74.60          \\
IMTCN            & 65.96          & 55.88          & 67.86          & 61.29          \\
SDTAN            & 80.14          & 73.33          & 78.57          & \textbf{75.86}          \\
FlightLLM-GLM      & 78.01          & 71.19          & 75.00          & 73.04          \\
FlightLLM-GPT      & \textbf{81.56} & 82.61          & 67.86          & 74.51          \\
FlightLLM-Deepseek & \textbf{81.56} & \textbf{85.71} & 64.29          & 73.47\\       \hline  
\end{tabular}
\end{table}

\subsection{Interpretability Analysis}
\begin{figure*}[htbp]
\centering
\captionsetup[subfigure]{skip=2pt} 
\begin{tabular}{@{}ccc@{}}
    \subfloat[IAS]{\includegraphics[width=0.31\textwidth]{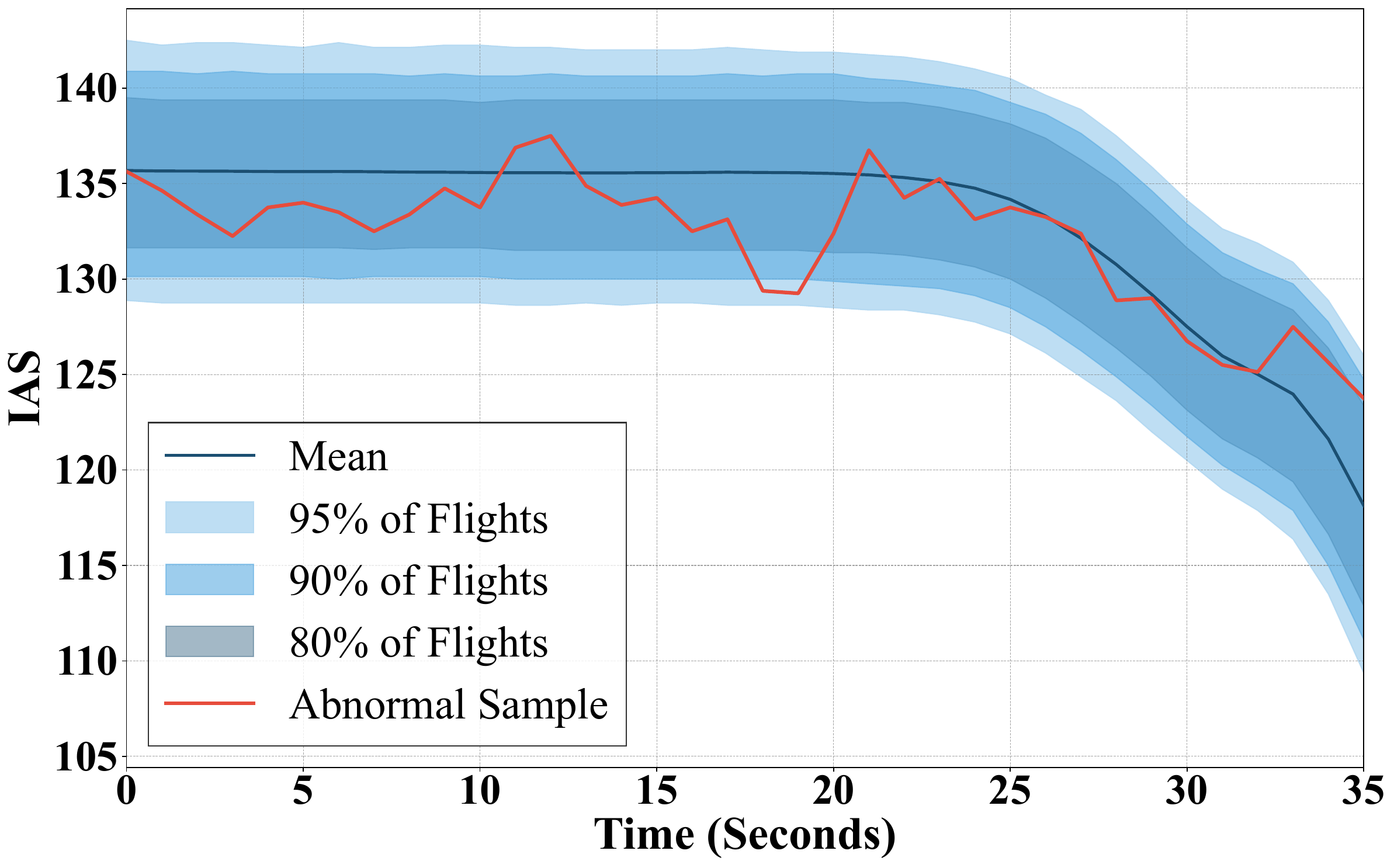}\label{IAS}} &
    \subfloat[ALT\_QNH]{\includegraphics[width=0.31\textwidth]{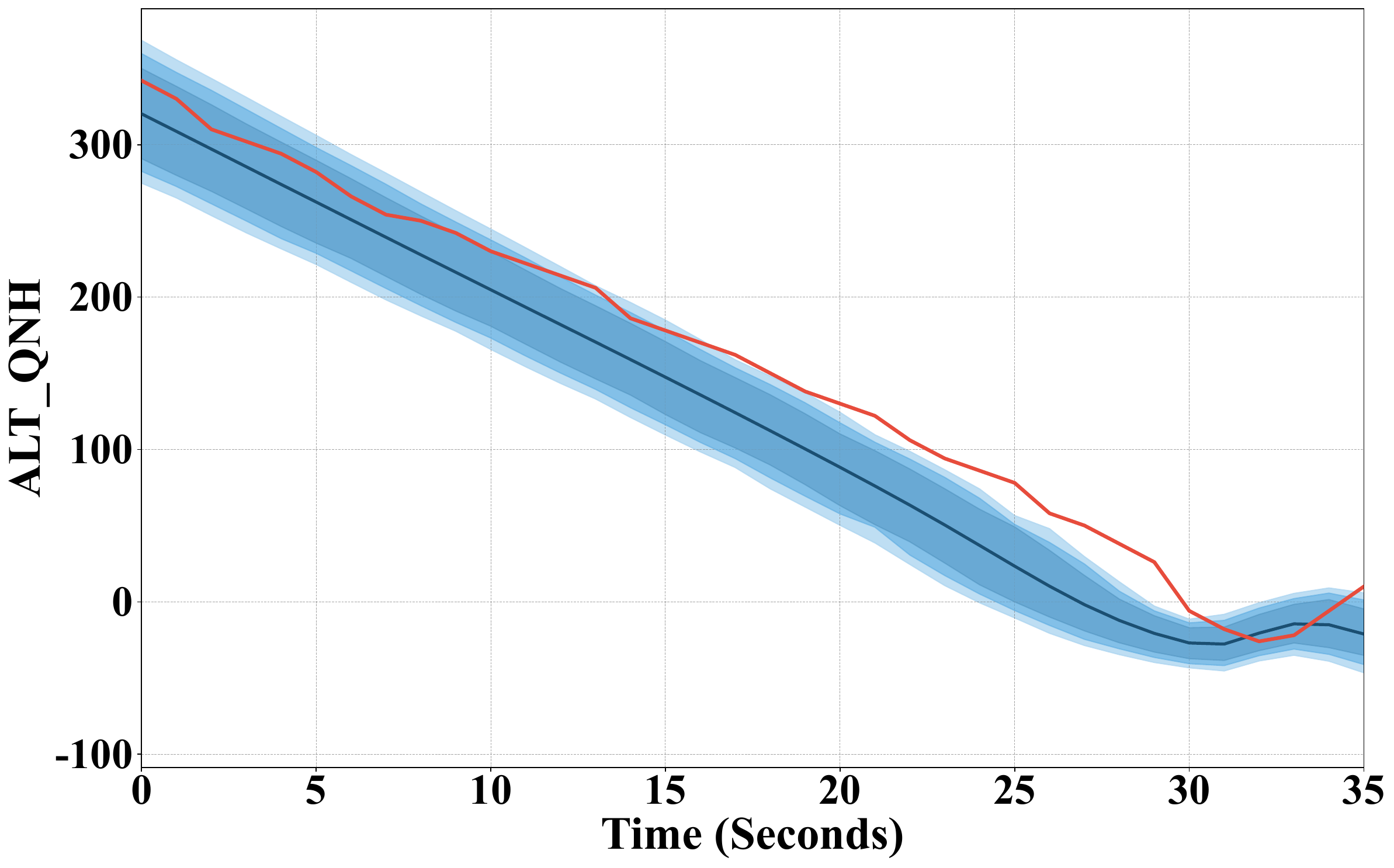}\label{ALT}} &
    \subfloat[WIN\_SPD]{\includegraphics[width=0.31\textwidth]{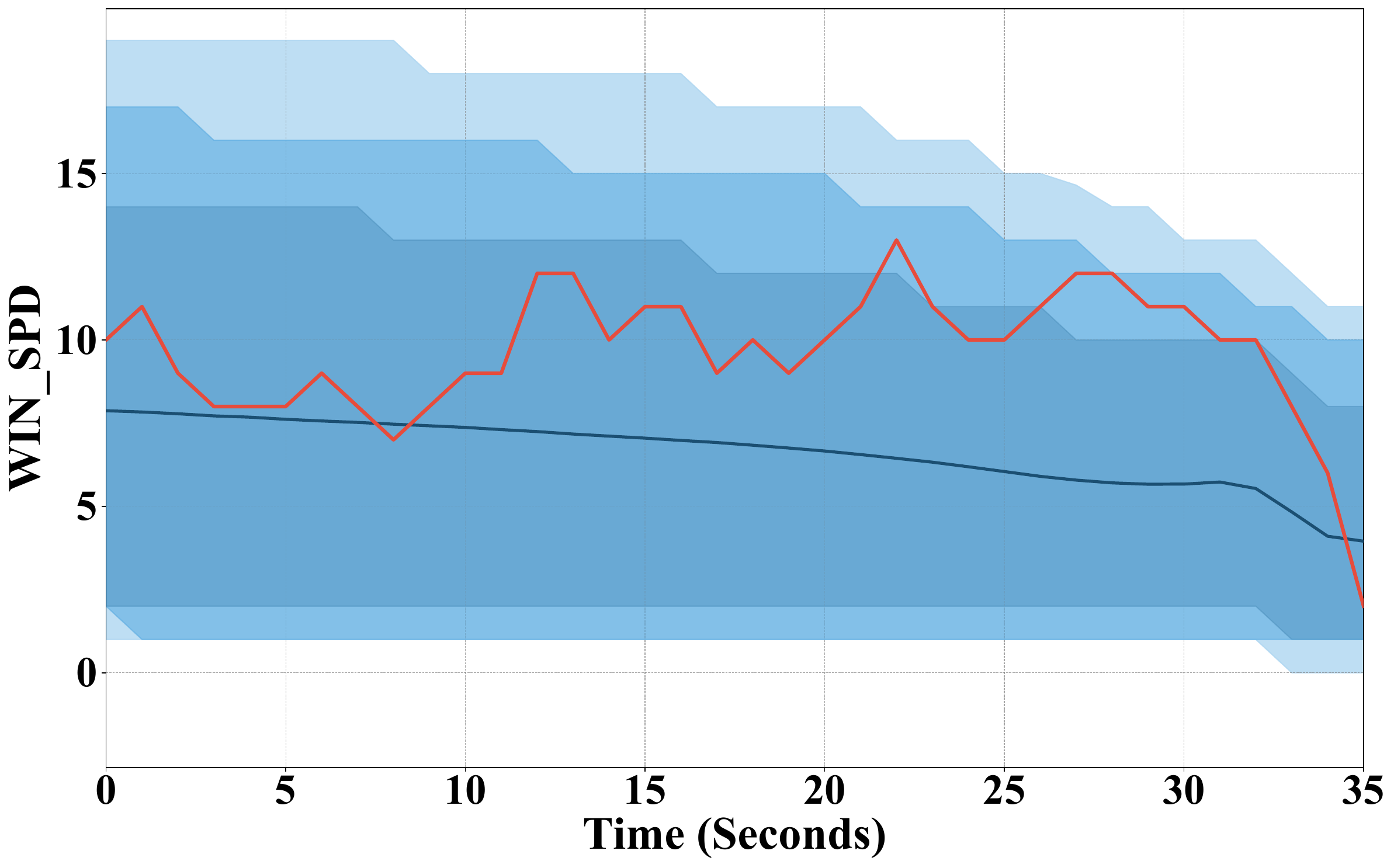}\label{WIN}} \\

    \subfloat[IVV]{\includegraphics[width=0.31\textwidth]{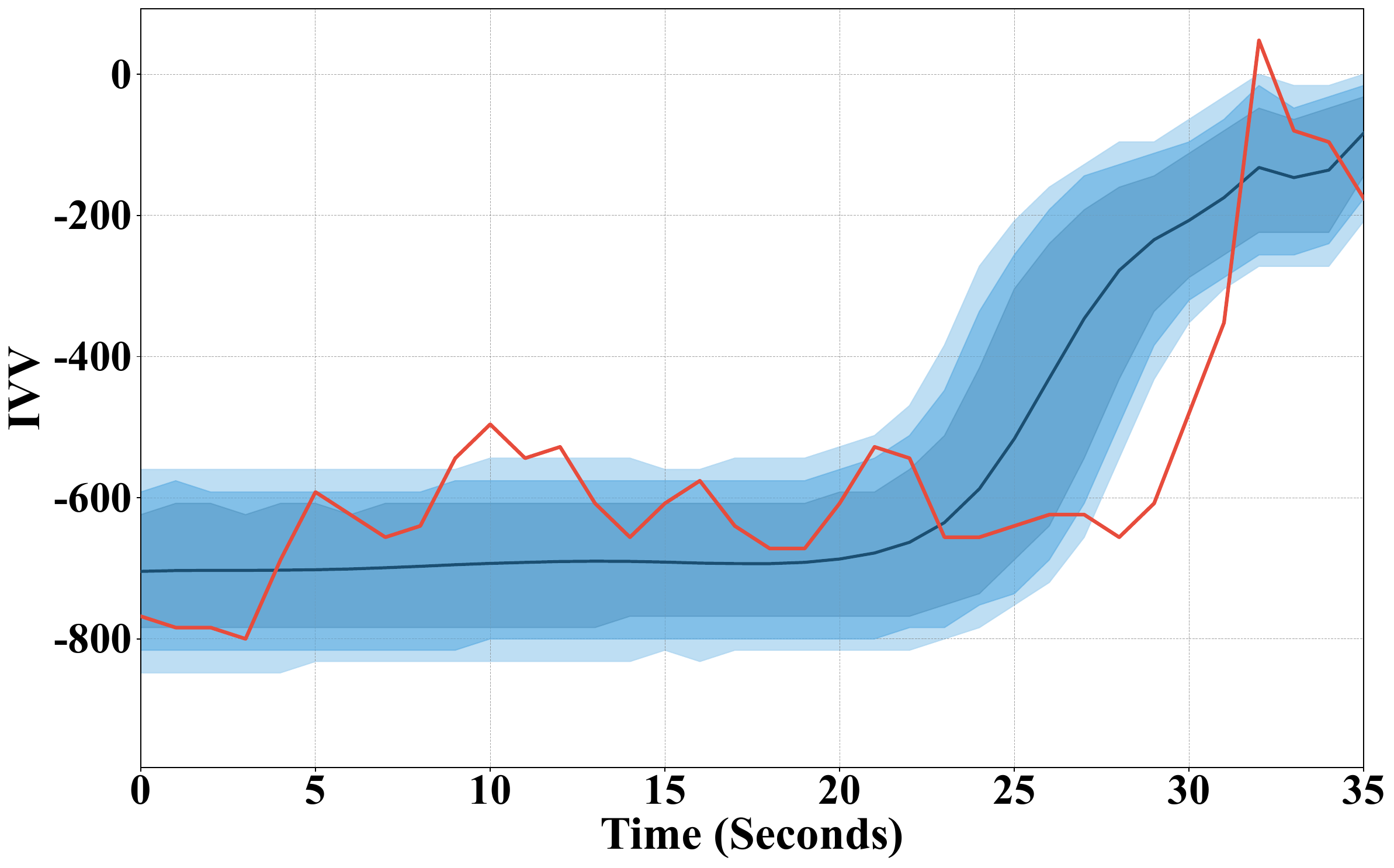}\label{IVV}} &
    \subfloat[PITCH]{\includegraphics[width=0.31\textwidth]{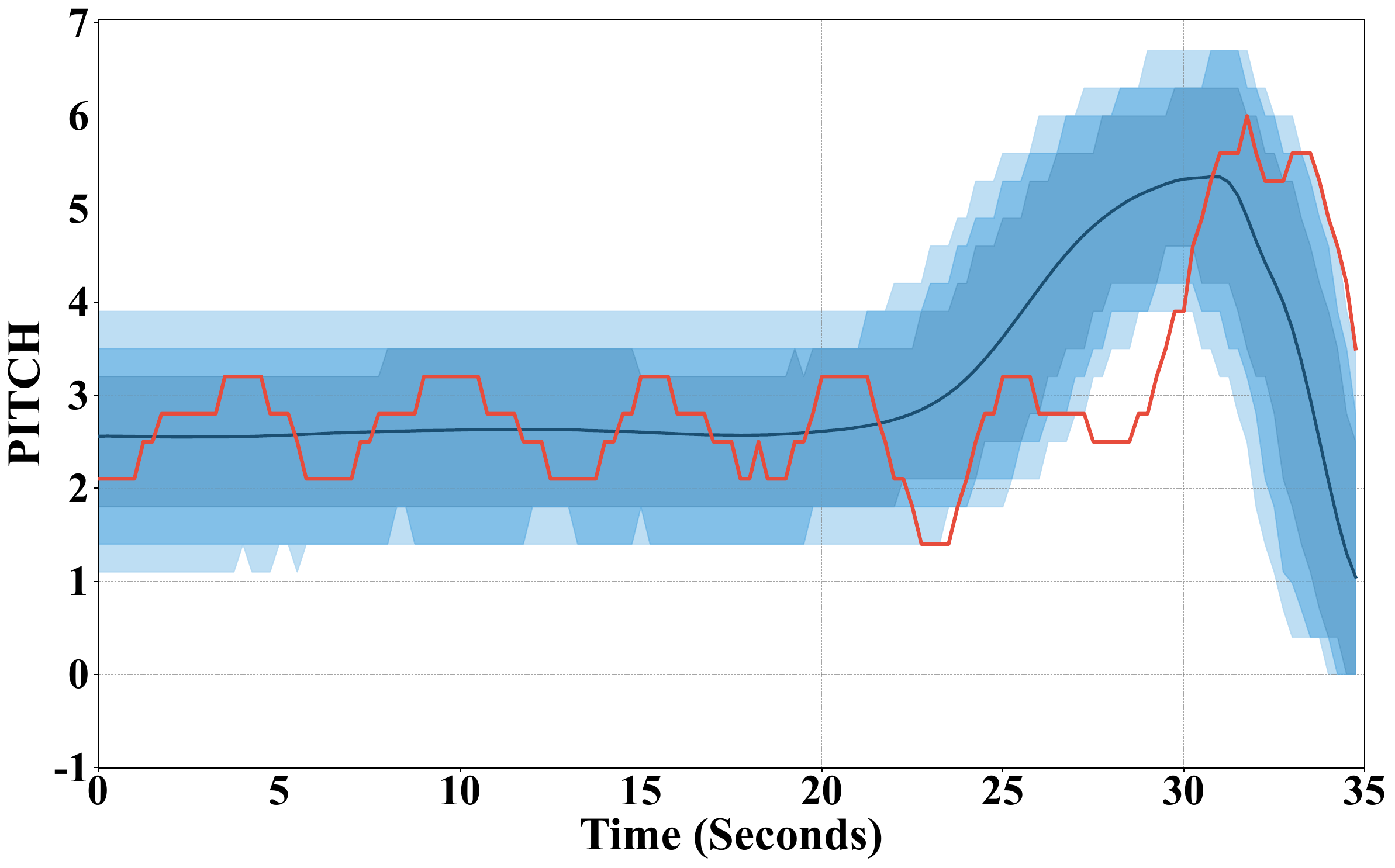}\label{PITCH}} &
    \subfloat[PITCH\_CMD]{\includegraphics[width=0.31\textwidth]{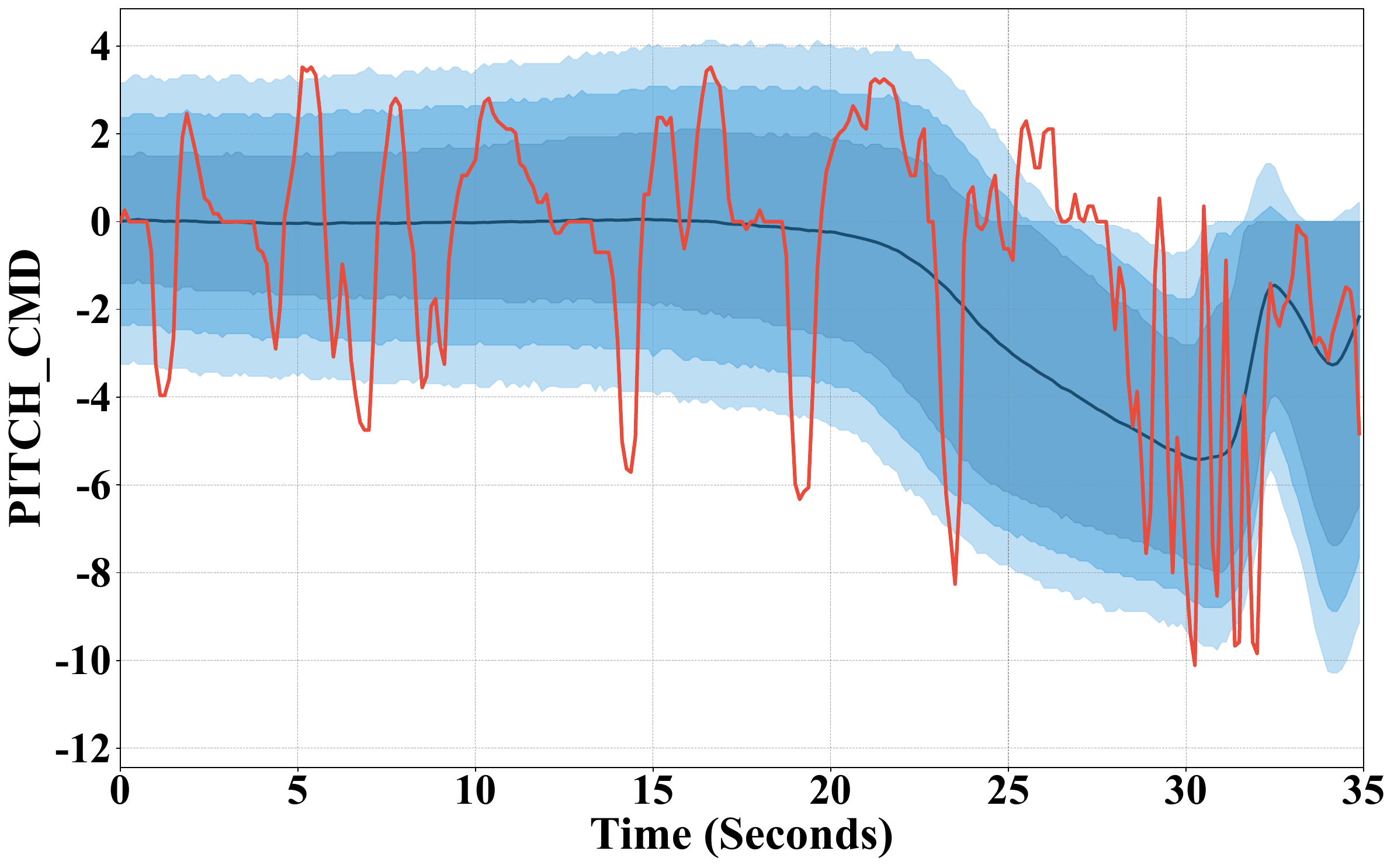}\label{PITCH_CMD}}
\end{tabular}
\caption{The parameters' curves of the analyzed case and corresponding curves of group.}
\label{curves}
\end{figure*}

To assess the attribution capability of the proposed FlightLLM in a practical setting, we conduct a case study on a representative hard landing sample. This sample is correctly classified as ``Hard Landing'' by the model.
Using the full dataset of normal landing samples from 37,929 real-world A320 flight records as a statistical reference, we compare the key flight parameters of this sample against the distributions derived from historical normal landings. This comparison allows us to quantify the deviation of the sample from standard patterns.
In addition, we incorporate the reasoning report generated by the FlightLLM to examine the causal logic behind its decision. By jointly analyzing statistical deviation and model-generated explanation, we evaluate whether the analysis from the LLM corresponds to real circumstances of this hard landing sample. The result demonstrates the output text is physically meaningful and consistent with the evidences from curves shown in Fig. \ref{curves}.

We first visualize the raw QAR trajectories of key flight parameters during the Final Approach Phase, as shown in Fig. \ref{curves}. The analysis focuses on deviations relative to the distribution of normal landing sample group.

According to Fig. \ref{curves}(a), at approximately 18\,s, the \textit{IAS} temporarily falls below the 10th percentile of the group, but it quickly recovers to the vicinity of the normal mean. This reflects instability within short term rather than sustained low-energy state.
During the 20-30s interval, the \textit{ALT\_QNH} of the hard landing flight exceeds the 95th percentile of normal samples, as shown in Fig. \ref{curves}(b). This indicates a high approach condition, suggesting the probability of abnormal flare operation.
In Fig. \ref{curves}(c), between 26\,s and 30\,s, \textit{WIN\_SPD} lies within the 80th–90th percentile range. Although not exceeding the extreme anomaly threshold, namely above the 95th percentile, the relatively strong and variable wind conditions likely increased operational workload during final approach.
As shown in Fig. \ref{curves}(d), the \textit{IVV} exceeds the 95th percentile at approximately 10\,s, indicating a descent steeper than normal flights. More critically, between 27\,s and 32\,s, the IVV curve rises sharply with a obvious gradient. This pattern suggests a rapid flare maneuver. And the timing of this flare is significantly later than that observed in 95\% of normal flights.
Fig. \ref{curves}(e) shows that the \textit{PITCH} trajectory corroborates this observation. Between 27\,s and 30\,s, the pitch angle remains below the 95\% distribution, reflecting an excessively flat attitude during the late approach phase. After 30\,s, the pitch angle increases abruptly with a steep slope, indicating a sharp pitch-up input. Similarly, the timing of this maneuver also lags behind the norm.
Furthermore, Fig. \ref{curves}(f) indicates \textit{PITCH\_CMD} exhibits continuous fluctuations throughout the approach, with severe oscillations immediately before and at touchdown. This behavior suggests intensive control inputs to correct pitch. Together, the delayed flare, oscillatory pitch commands, and relatively unstable wind ultimately contribute to the hard landing outcome.

\begin{figure*}[htbp]
\centering
\includegraphics[scale=0.17]{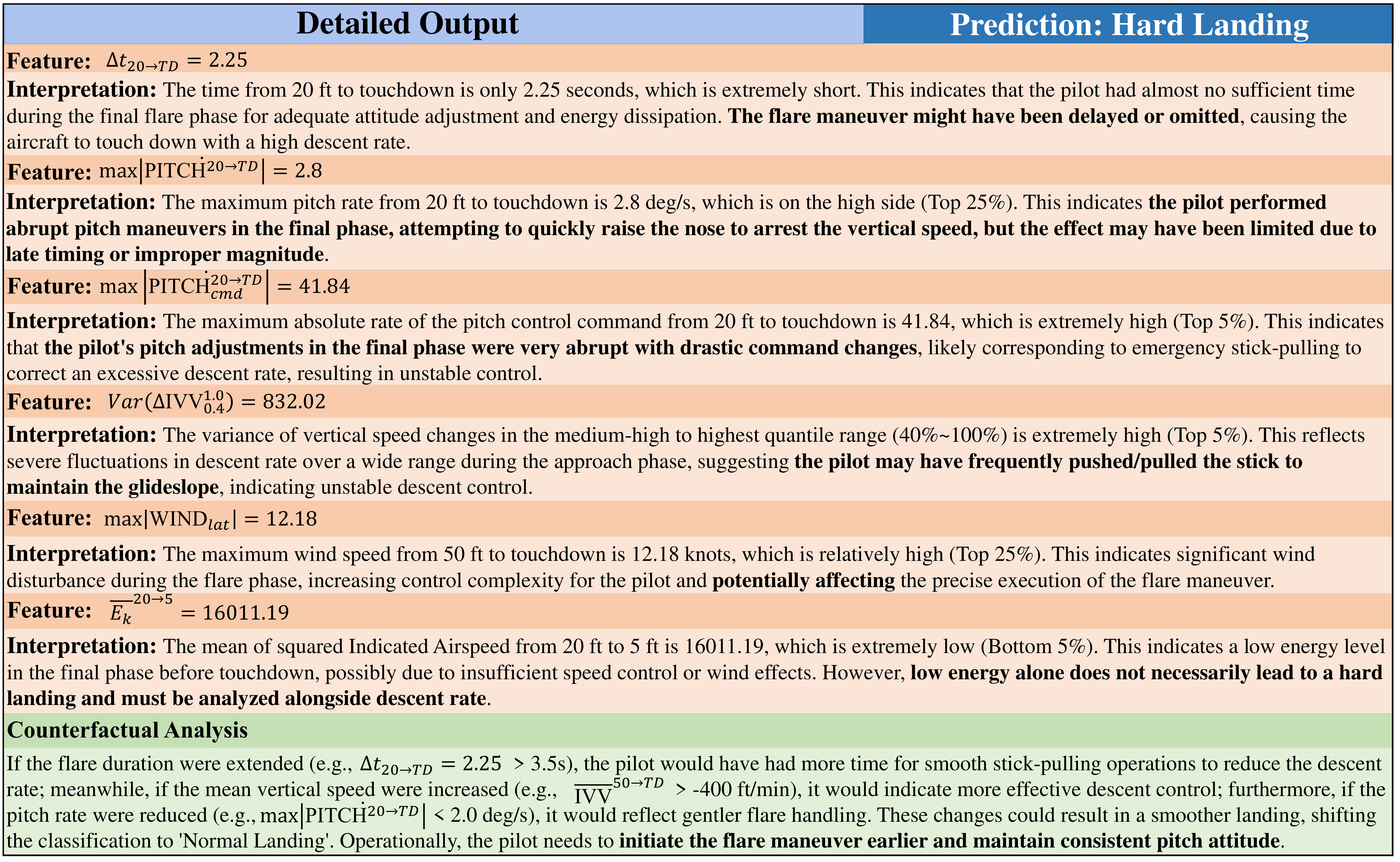}
\caption{The detailed textual output of LLM.}
\label{llmout}
\end{figure*}

FlightLLM identifies the physical patterns mentioned above and generates a reasoning chain that aligns closely with the visual analyses of the flight parameter trajectories. The output explanation can capture the delayed flare timing and rapid pitch correction and also distinguish between the main factors and the secondary factors, demonstrating consistency between attribution from LLMs and the physical evidence shown in Fig. \ref{curves}. The main analysis from LLMs is as follows:
\begin{itemize}
    \item Identification of Late Flare: For the features $\Delta t_{20\to TD}$ and $\max|\dot{\text{PITCH}}^{20\to TD}|$, the LLM provided the following interpretations: ``The flare maneuver might have been delayed or omitted'' and ``the pilot performed abrupt pitch maneuvers in the final phase, attempting to quickly raise the nose to arrest the vertical speed, but the effect may have been limited due to late timing or improper magnitude.'' These statements are consistent with the trajectory analysis presented earlier. In particular, the delayed pitch-up timing and the sharp gradient observed in Fig. \ref{curves}(d) and Fig. \ref{curves}(e) curves support the model’s conclusion that the flare was initiated late and executed aggressively. 
    \item Identification of Drastic Correction: For the features $\max|\dot{\text{PITCH}}_{cmd}^{20\to TD}|$ and $Var(\Delta \text{IVV}_{0.4}^{1.0})$, the LLM generated the following interpretations: ``pilot's pitch adjustments in the final phase were very abrupt with drastic command changes'' and ``the pilot may have frequently pushed/pulled the stick to maintain the glideslope.'' These interpretations are strongly consistent with severe fluctuations observed in Fig. \ref{curves}(f). 
    \item Prioritization: Although both \textit{WIN\_SPD} and \textit{IAS} show statistical deviations from the group distribution, the model did not overly emphasize these factors as primary causes. Instead, the LLM identified these environmental variables as secondary contributors and assigned greater weight to features related to pilot control actions such as pitch and vertical speed. Rather than regarding each statistical abnormality at the same level, the model differentiates between background environmental conditions and decisive operational inputs, focusing on pilot control behaviors that directly influence the landing outcome. 
    \item Actionable guidance: Through counterfactual reasoning, the model generated explicit operational recommendations. For example, it stated that pilots should ``initiate the flare earlier and maintain a consistent pitch attitude.'' This form of output extends beyond diagnosis post event. By translating attribution results into concrete suggestions, the approach narrows the gap between theoretical analysis and actionable training.
\end{itemize}
The accordance between reasoning at feature level and raw QAR data evidence further validates the attribution capability of the FlightLLM. The complete textual output from LLM is shown in Fig. \ref{llmout}.

\subsection{Ablation Experiment}
To thoroughly evaluate the individual contributions of the core components within the FlightLLM, we design three model variants for ablation analysis. Each variant respectively removes a specific module and keeps the remaining components unchanged. 
The comparison results of the ablation experiment are summarized in Table \ref{ablation}.

Ablation results provide the evidence of the contribution of each module. \textbf{(\romannumeral1) w/o Semanticization: } Semantic Discretization module is removed in Variant A. Compared with the full FlightLLM, removing the semantic discretization module causes Precision to drop sharply from 0.8571 to 0.6716. This decline indicates that when exposed directly to raw number inputs, the LLM performs worse in forming stable decision boundaries. In such cases, when the model deals with features with large absolute values, it tends to treat them as abnormal signals. Consequently, this leads to high Recall but an inflated False Positive Rate.  This is a sound proof that semantic discretization solves this issue by converting continuous numerical values into qualitative labels. These labels act as cognitive anchors, allowing the LLM to reason within an explicitly defined semantic scale rather than relying on implicit number comparison. As a result, classification Precision improves substantially. 
\textbf{(\romannumeral2) w/o Expert Hint: } Statistical Expert Hinting mudule is removed in Variant B. In this condition, the model’s Accuracy decreases to 0.7518, and the F1-Score becomes the lowest among all variants. This outcome demonstrates the importance of prior probabilistic guidance. The prediction probabilities generated by CatBoost constrain the reasoning space of the LLM. Without this guidance, the model is more susceptible to ambiguous features and may produce unstable judgments, leading to more conservative predictions, reducing Recall and degrading overall balance between Precision and Recall.
\textbf{(\romannumeral3) w/o Context Retrieval: } Dynamic Context Retrieval module is removed in Variant C, namely Zero-Shot inference. The variant achieves the lowest Accuracy among all variants. This result shows the limitation of Zero-Shot reasoning in complex safety scenarios. Without contrastive samples, the LLM lacks context and may make an overly aggressive decision. It tends to interpret minor fluctuations as decisive abnormal indicators. 
By incorporating dynamically retrieved positive and negative samples, the LLM can compare subtle differences between similar normal and abnormal samples, leading to a more clear decision boundary. Consequently, FlightLLM achieves better Accuracy and Precision. 

Notably, experimental results show that Semantic Discretization module and Dynamic Context Retrieval module significantly improve classification accuracy, but they also lead to a certain degree of cognitive convergence.
For Semantic Discretization module, borderline high-risk samples may be assigned to less severe semantic categories because their values fall just below a predefined threshold. For example, a feature closed to but lower than the ``Extremely High” boundary may be labeled as ``Slightly High.” In such cases, the LLM may confidently regard this sample as not in danger, leading to missed detections.
As for Dynamic Context Retrieval, the model becomes more conservative when evaluating previously unseen borderline hazardous patterns, which leads to lower recall. 
Since the objective of our study is to provide high-quality diagnostic explanations for flight safety specialists rather than act as a simple alarm trigger, this trade-off is acceptable. A moderate reduction in recall is exchanged for better accuracy and interpretability, ultimately reducing the burden on experts.

\begin{table}[]
\caption{The result of ablation experiment}
\label{ablation}
\setlength{\tabcolsep}{4pt} 
\begin{tabular}{lcccc}
\hline
Variant                        & Accuracy       & Precision      & Recall         & F1             \\ \hline
Variant A(w/o Semanticization) & 76.60          & 67.16          & 80.36          & 73.17          \\
Variant B(w/o Expert Hint)     & 75.18          & 75.61          & 55.36          & 63.92          \\
Variant C(w/o Context Retrieval)        & 73.76          & 63.01          & \textbf{82.14} & 71.32          \\
FlightLLM                        & \textbf{81.56} & \textbf{85.71} & 64.29          & \textbf{73.41} \\ \hline
\end{tabular}
\end{table}

\subsection{Consistency Analysis}
\begin{figure}
    \includegraphics[scale=0.25]{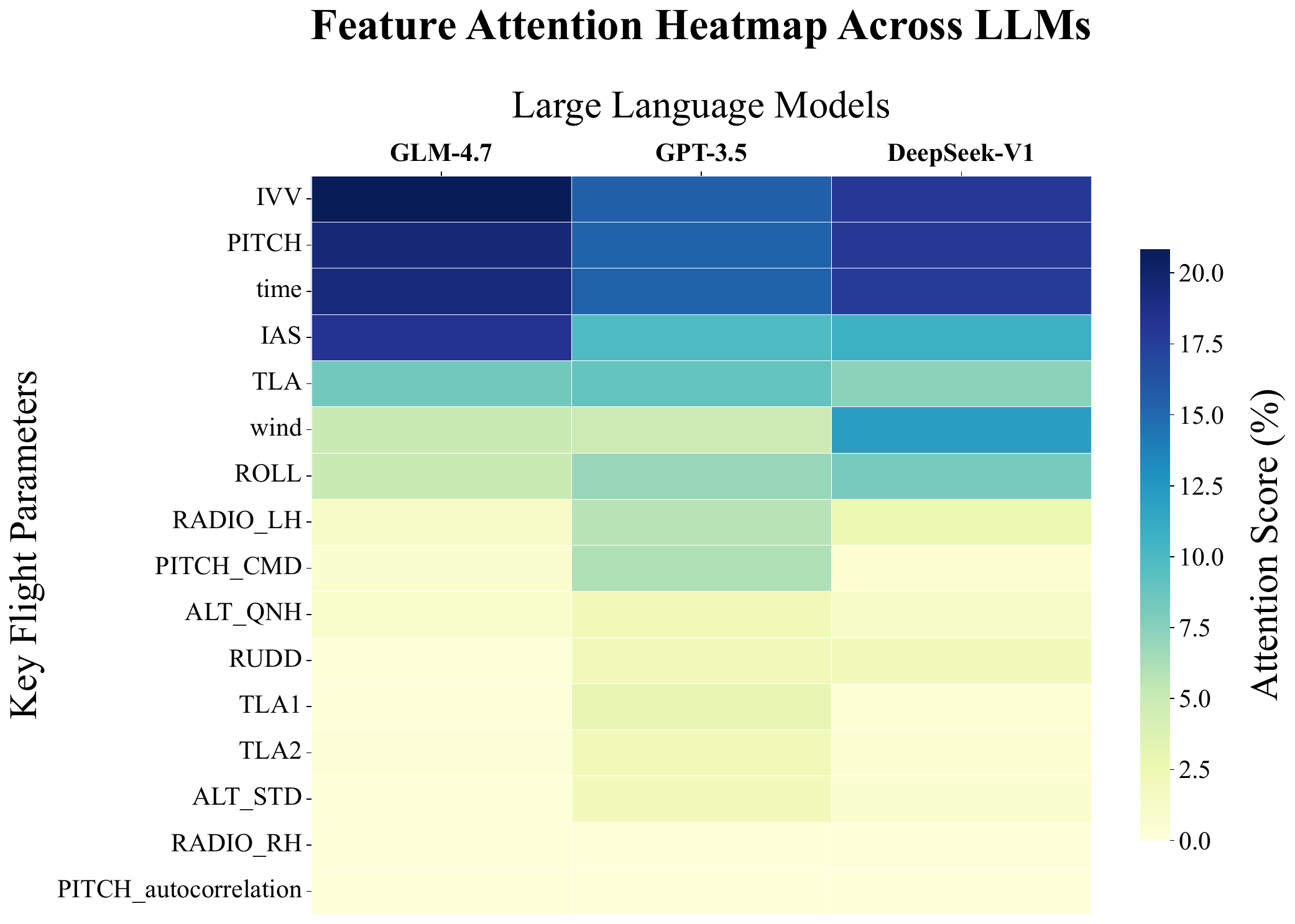}
    \caption{Feature Attention Heatmap of 3 language models. }
    \label{heatmap}
\end{figure}

To examine attribution consistency among different models, we visualize the feature distributions of the three LLMs in Fig.~\ref{heatmap}. The heatmap illustrates the degree of alignment in feature-level across models.
At the top of the heatmap, all three models consistently assign the highest importance to \textit{IVV}, \textit{PITCH}, and time. These directly correspond to the three primary causes of hard landing: abnormal vertical speed, pitch angle, and flare time. This convergence suggests that the classification results are not random. Instead, the models capture the core mechanisms underlying hard landing events. Regardless of the backbone model, the primary factors remain stable.
However, the models exhibit differences when evaluating secondary factors. GLM-4.7 exhibits a relatively high sensitivity to \textit{IAS}, indicating that it assigns high weight to airspeed. While airspeed anomaly may signal instability due to wind or throttle adjustments, they are typically indirect contributors rather than dominant causes. Over emphasis on such secondary indicators may partially explain the slightly lower performance of GLM-4.7 compared to the other models.
GPT-3.5 demonstrates a more distributed attribution pattern. It allocates moderate weights across a wider set of variables, resulting in a broader but less concentrated reasoning structure. This behavior reflects a tendency to incorporate additional contextual cues into the explanation process.
Overall, the heatmap analysis reveals strong agreement on primary causal factors and subtle divergence in some feature weighting, providing further evidence that FlightLLM can activate reasoning consistent with flight mechanism across different LLM backbones.

\section{Conclusion}
In this paper, we propose FlightLLM for interpretable flight safety event analysis. The approach bridges the gap between QAR time series data and text data needed by LLMs, compensates for the shortage of limited data and injects domain knowledge into LLMs, enabling them to perform better in classification, and more importantly, generate causal explanations for flight safety events represented by hard landing.
The Feature Engineering module converts raw QAR data into hybrid feature vectors that combine statistical descriptors with physically meaningful indicators. 
The Semantic Discretization module maps continuous numerical values to qualitative semantic labels. It addresses the known limitations of LLMs in numerical token processing to some extent. 
The Statistical Expert Hinting module introduces a traditional model as an assistant, providing an anchor for LLMs to infer.
In addition, the Dynamic Context Retrieval module selects contrastive samples, providing contextual information that stabilizes classification decisions.
Experimental results indicate that FlightLLM achieves competitive classification performance while offering explanations consistent with aviation dynamics. The approach shows unprecedented interpretability and provides actionable advice, which is significant for improving flight safety. 

Despite the encouraging results, our work also has several limitations. 
First, we did not fine-tune the large language models specific to hard landing. All experiments were conducted using publicly available and pre-trained models. Although this approach has the generalization capability, fine-tuning aimed at aviation field may further enhance reasoning consistency and performance.
Second, although FlightLLM can be extended to analyze multiple flight safety events, modifying the prompts is required to ensure consistency with the underlying physical mechanisms of each event. This event-specific prompt modification introduces additional manual effort.
Future research can therefore focus on developing more transferable prompt engineering strategies. A unified and physically grounded prompting framework would enhance scalability and robustness for diverse aviation safety scenarios.

\bibliographystyle{IEEEtran}
\bibliography{IEEEabrv,ref}

\begin{IEEEbiography}[{\includegraphics[width=1in,height=1.25in,clip,keepaspectratio]{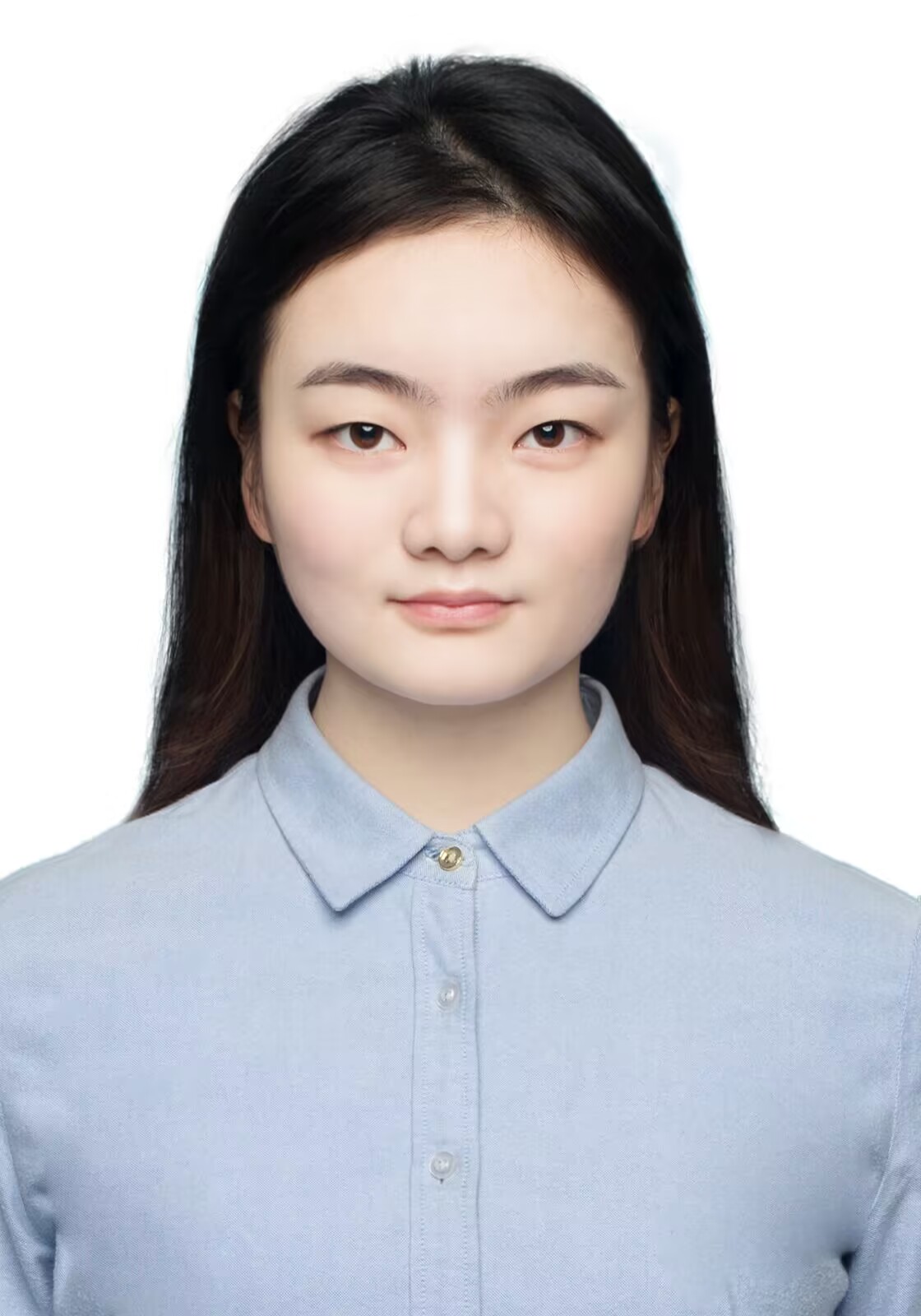}}]{Lu Xu} was born in Bijie, Guizhou, China, in 2002. She received the B.S. degree in Computer Science and Technology from Wuhan University, Wuhan, China, in 2025. She is currently pursuing the master’s degree with the College of Computer Science, Chongqing University. Her research interests include data analysis, explainable artificial intelligence, and flight safety. 
\end{IEEEbiography}

\begin{IEEEbiography}[{\includegraphics[width=1in,height=1.25in,clip,keepaspectratio]{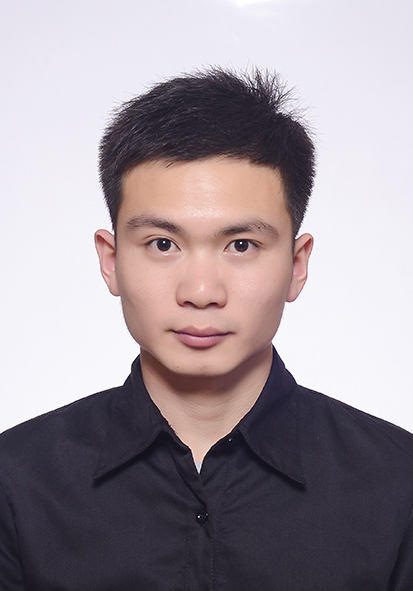}}]{Xu Li} was born in Baoying, Jiangsu, China, in 1994. He received the B.S. degree in Computer Science and Technology from Nanjing Forestry University, Nanjing, China, in 2018 and the M.S. degree in Computer Science and Technology from Chongqing University, Chongqing, China, in 2021. Currently, he is a PhD candidate in the college of computer science, Chongqing University. His research interests include flight data analysis, explainable artificial intelligence, and data mining. He has published high quality papers in TITS, KBS, UIC, etc.
\end{IEEEbiography}

\begin{IEEEbiography}[{\includegraphics[width=1in,height=1.25in,clip,keepaspectratio]{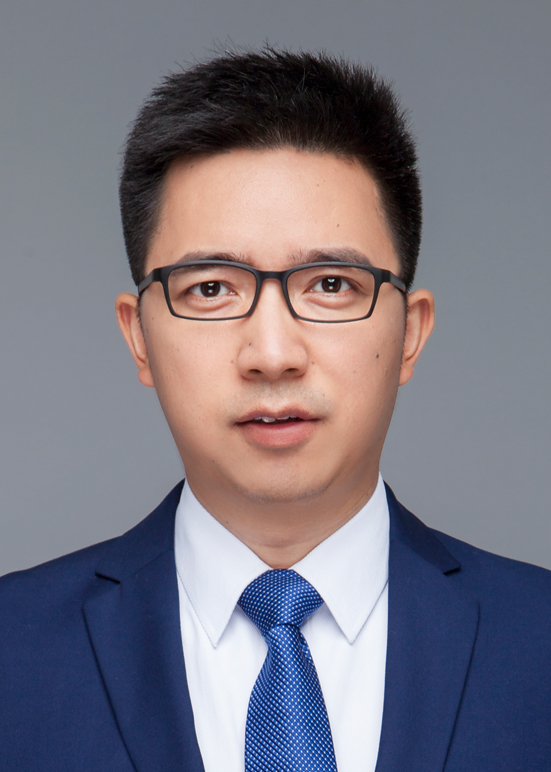}}]{Linjiang Zheng}(Member, IEEE) was born in Linshui, Sichuan, China in 1983. He received the Ph.D. degree in Computer Science and Technology from Chongqing University, China, in 2010. He currently works as a professor at the college of computer science in Chongqing University, China. His research interests include internet of things, transportation big data, RFID application, etc. He has published 50+ high quality journal and conference articles, including TKDE, TNNLS, TITS, TVT, etc.
\end{IEEEbiography}

\begin{IEEEbiography}[{\includegraphics[width=1in,height=1.25in,clip,keepaspectratio]{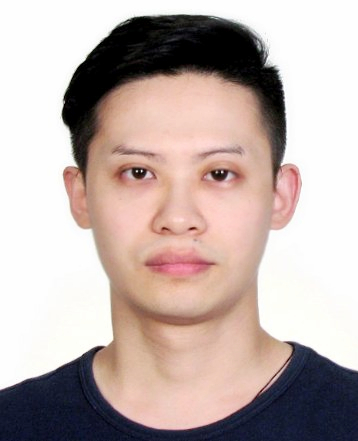}}]{Fan Li} received his M.Sc. and Ph.D. degrees in mathematical statistics from Sichuan University, China, in 2009 and 2012, respectively. In 2012, he joined as a faculty member of Sichuan University, where he was also a postdoctoral researcher with the Laboratory of Prognostics and Health Management until July 2019. He is currently an associate researcher with the Key Laboratory of Flight Techniques and Flight Safety, CAAC, Civil Aviation Flight University of China. His research interests are mainly concerned with the application of big data analytics and artificial intelligence to intelligent civil aviation, including deep learning-based anomaly detection and precursor detection and multisensor tracking fusion.
\end{IEEEbiography}

\begin{IEEEbiography}[{\includegraphics[width=1in,height=1.25in,clip,keepaspectratio]{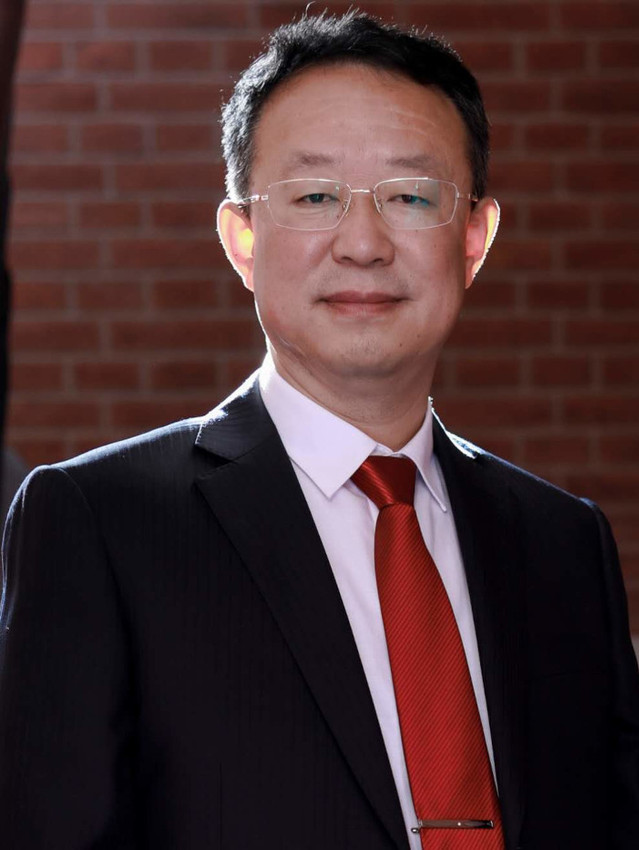}}]{Riquan Zhang} received the Ph.D. degree in Probability Theory and Mathematical Statistics, East China Normal University, Shanghai, China, in 2003. He currently works as a professor and dean at the School of Statistics and Data Science, Shanghai University of International Business and Economics, Shanghai, China. His research interests include big data statistics, statistical machine learning, financial statistics, etc. He is the author of 4 monographs and 3 textbooks, including 2 national textbooks. He has published 230+ high papers in prestigious journals and conferences. He is the PI and Co-PI of 20+ projects, including National Natural Science Foundation of China (NSFC), provincial and ministerial key projects.
\end{IEEEbiography}

\begin{IEEEbiography}[{\includegraphics[width=1in,height=1.25in,clip,keepaspectratio]{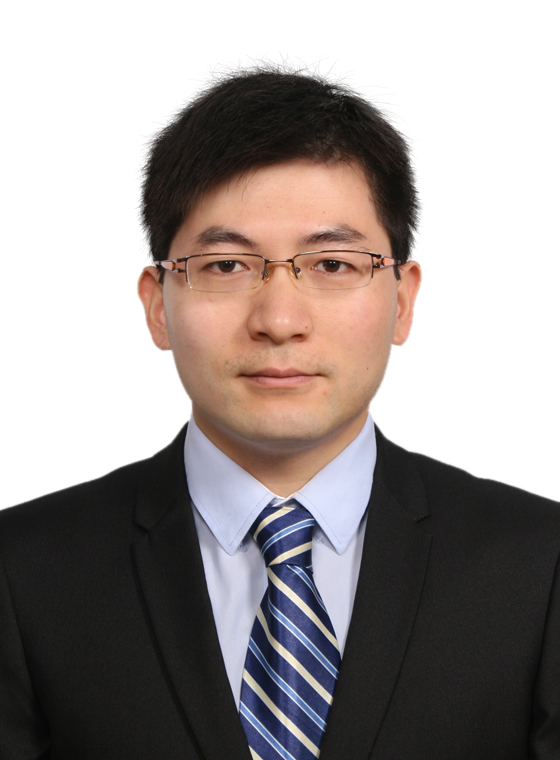}}]{Jiaxing Shang}(Member, IEEE) received the B.S. and Ph.D. degrees in Control Science and Engineering from Tsinghua University, Beijing, China, in 2010 and 2016 respectively. Currently, he is a professor at the College of Computer Science in Chongqing University, Chongqing, China and a Marie Sklodowska-Curie Postdoctoral Fellow with the University of Exeter, Exeter, UK. His research interests include industrial big data mining, explainable AI, social network analysis and mining, etc. He has published 100+ high quality journal and conference articles, including TKDE, TNNLS, TITS, TETC, WSDM, etc.
\end{IEEEbiography}

\end{document}